\documentclass[journal,trans]{IEEEtran}
\usepackage{lineno}
\usepackage{colortbl}
\usepackage{graphicx, color}
\usepackage{tikz}
\usepackage{float}
\usepackage{subcaption}
\usepackage{epstopdf}
\usepackage{algpseudocode,algorithm,algorithmicx}
\usepackage{verbatim}
\usepackage{url}
\usepackage{amsfonts}
\usepackage{graphics} 
\usepackage{epsfig} 
\usepackage{amsmath}

\modulolinenumbers[5]

\def\l|{\left\|}
\def\r|{\right\|}

\def\bfc{{\mbox{\boldmath $c$}}}
\def\bfd{{\mbox{\boldmath $d$}}}

\def\bfp{{\mbox{\boldmath $p$}}}

\def\bfu{{\mbox{\boldmath $u$}}}
\def\bfv{{\mbox{\boldmath $v$}}}

\def\bfx{{\mbox{\boldmath $x$}}}

\def\bfz{{\mbox{\boldmath $z$}}}

\def\bfD{{\mbox{\boldmath $D$}}}

\def\bfF{{\mbox{\boldmath $F$}}}
\def\bfG{{\mbox{\boldmath $G$}}}

\def\bfI{{\mbox{\boldmath $I$}}}

\def\bfM{{\mbox{\boldmath $M$}}}

\def\bfR{{\mbox{\boldmath $R$}}}

\def\bfU{{\mbox{\boldmath $U$}}}
\def\bfV{{\mbox{\boldmath $V$}}}

\def\bfX{{\mbox{\boldmath $X$}}}
\def\bfY{{\mbox{\boldmath $Y$}}}

\def\bfLambda{{\mbox{\boldmath $\Lambda$}}}

\def\bfSigma{{\mbox{\boldmath $\Sigma$}}}

\ifCLASSINFOpdf

\begin{document}
\title{A Transfer Learning Approach to Cross-modal Object Recognition: from Visual Observation to Robotic Haptic Exploration}


\author{\IEEEauthorblockN{Pietro Falco \IEEEauthorrefmark{1}\IEEEauthorrefmark{2}~\IEEEmembership{Member IEEE},
Shuang Lu\IEEEauthorrefmark{1},
Ciro Natale\IEEEauthorrefmark{3},
Salvatore Pirozzi\IEEEauthorrefmark{3},\\
and
Dongheui Lee\IEEEauthorrefmark{1}\IEEEauthorrefmark{4}~\IEEEmembership{Member IEEE}} \\
\IEEEauthorblockA{\IEEEauthorrefmark{1}Technical University of Munich, Germany}\\
\IEEEauthorblockA{\IEEEauthorrefmark{2}ABB Corporate Research, Sweden}\\
\IEEEauthorblockA{\IEEEauthorrefmark{3}University of Campania Luigi Vanvitelli, Italy}\\
\IEEEauthorblockA{\IEEEauthorrefmark{4}Institute of Robotics and Mechatronics, German Aerospace Center, Germany}

\thanks{
Corresponding author: P. Falco (email: pietro.falco@se.abb.com).}}

\markboth{IEEE Transactions on Robotics}%
{Shell \MakeLowercase{\textit{et al.}}: Bare Demo of IEEEtran.cls for IEEE Transactions on Magnetics Journals}
%

\IEEEtitleabstractindextext{%
\begin{abstract}
In this work, we introduce the problem of cross-modal visuo-tactile object recognition with robotic active exploration.
With this term, we mean that the robot observes a set of objects with visual perception and, later on, it is able to recognize such objects only with tactile exploration, without having touched any object before.
Using a machine learning terminology, in our application we have a visual training set and a tactile test set, or vice versa.
To tackle this problem, we propose an approach constituted by {four} steps: finding a visuo-tactile common representation, defining a suitable set of features, transferring the features across the domains, and classifying the objects.
We show the results of our approach using a set of $15$ objects, collecting $40$ visual examples and {five} tactile examples for each object.
The proposed approach achieves an accuracy of $94.7\%$, which is comparable with the accuracy of the monomodal case, i.e., when using visual data both as training set and test set. Moreover, it performs well compared to the human ability, which we have roughly estimated carrying out an experiment with {ten} participants.
\end{abstract}

}

\maketitle

\IEEEdisplaynontitleabstractindextext
%
\IEEEpeerreviewmaketitle

\section{INTRODUCTION} \label{s:intro}
Multi-modal perception technologies are the key enablers of robot
autonomy operating in unstructured environments. On one hand,
{computer} vision is fundamental for scene analysis and motion
planning of the robot or for monitoring the robot workspace. On the
other hand, vision cannot be the only solution to the perceptual
need of a robot autonomously interacting with an unknown
environment. In fact, effectiveness of a visual system is affected by
lighting conditions, occlusions and limited field of view,
especially during physical interaction with the world. Therefore,
vision has to be supported by additional perceptual abilities, such
as force and tactile feedback, that is extremely rich of useful
information when the robot is in contact with the environment, e.g.,
measuring the contact force allows the robot to immediately identify
constrained directions where the motion is not allowed.
These considerations motivated a lot of research effort in the last
decade towards advancements in the multi-modal perception
technology, especially in the combined use of vision and
force/tactile sensing. However, the field of cross-modal perception
has been explored only superficially, while leveraging knowledge
acquired in a perceptual domain during the execution of an action exploiting
a different sensing modality could lead to a great level of
autonomy. The present paper focuses on a typical application of
cross-modal perception, namely the visuo-tactile object recognition,
which means recognizing a previously seen object (never touched
before) by simply touching it.

{In neuroscience and psychology, cross-modal (or intermodal) object
recognition is defined as \emph{the name for the ability to
recognize an object, previously inspected with one modality like
vision, via a second modality like touch \rm{\cite{dictionary}}},
\emph{without prior training in the second modality
\rm{\cite{schumacher2016cross}}.}
\begin{figure}[t]
\centering
     \includegraphics[width=\columnwidth]{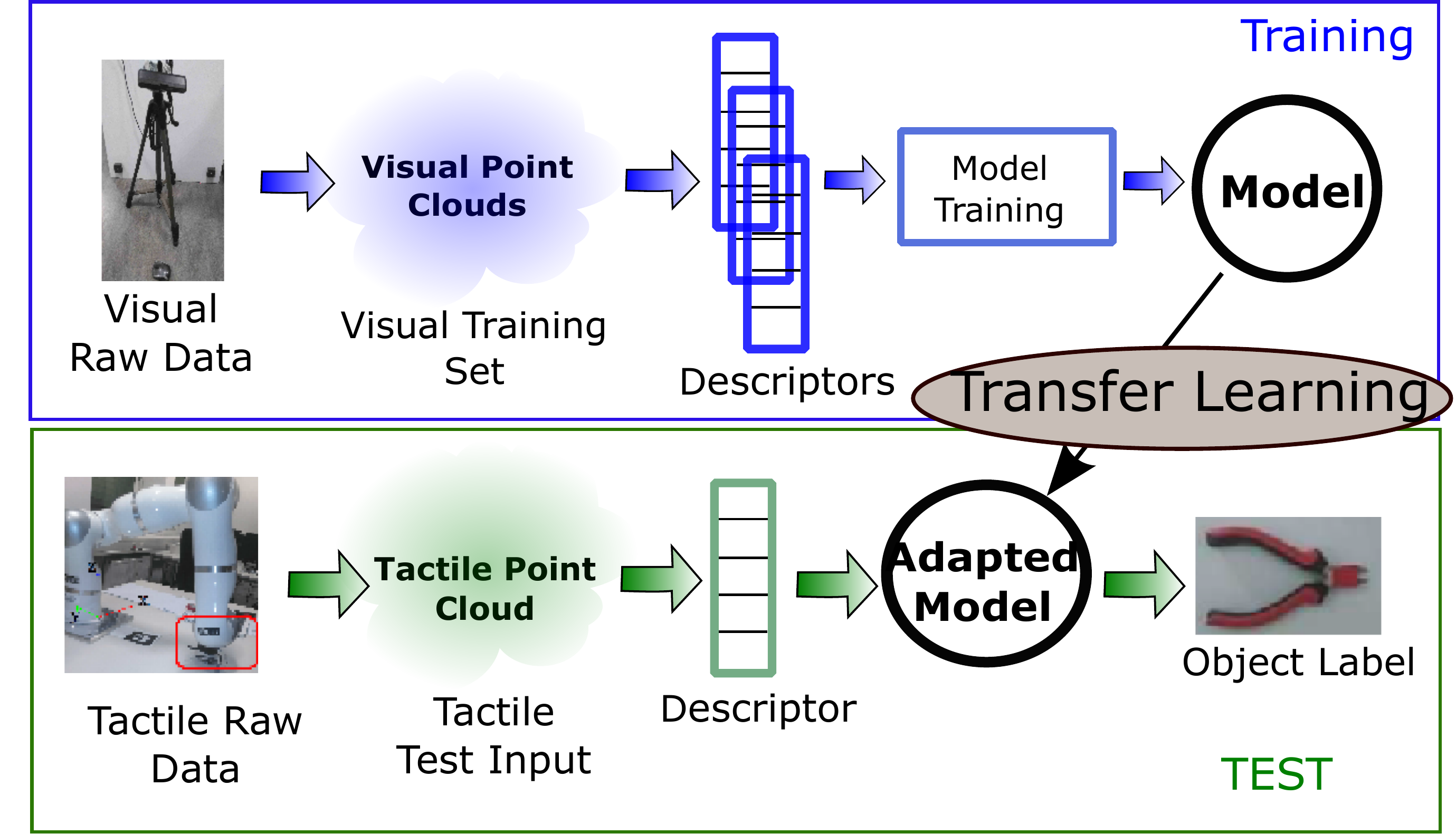}
    \caption{Cross-modal recognition concept: training pipeline (top) and execution pipeline (bottom)}
    \label{fig:concept}
\end{figure}
{Cross-modal perceptual ability could enhance autonomy of a robot that is performing a task exploiting a given mono-modal
perception system. The used sensor may unexpectedly become unavailable, e.g,
vision due to lighting failure or occlusion, and we explored if the robot is able to perform its task using another sensor modality, e.g, a tactile sensor.
To achieve this objective, the robot must be able to exploit its
a-priori knowledge gained in the first sensing modality and use it
at run time exploiting the second sensing modality.
The concept of this approach is depicted in Fig.~\ref{fig:concept}.
We investigate if it is possible to recognize an object by using
only tactile data and a classifier trained only with visual data.}

{In order to make a cross-modal object recognition algorithm effective, two research challenges have to be faced, i.e.,}
\begin{enumerate}
\item Selecting a common representation for both visual and tactile
data. It should be fully interchangeable in a transparent way in
both domains.
\item Selecting the right descriptors for the chosen visuo-tactile
representation, possibly by re-using descriptors from the computer
vision community.
\end{enumerate}
This paper presents how we tackle these challenges by
supporting our proposed choices with a large set of experiments.
The preliminary version of the proposed approach is presented in \cite{falco2017cross} and extended by
leveraging results from transfer learning \cite{pan2010survey} to
further improve the performance. With respect
to the work in \cite{falco2017cross}, the novel additional research
questions are
\begin{itemize}
  \item do transfer learning approaches help to improve the performance of cross-modal visuo-tactile problems?
  \item are transfer learning approaches alternative to the solution proposed in \cite{falco2017cross} or can they be effectively combined?
\end{itemize}
The rest of the paper is organized as follows.
Section~\ref{s:related} discusses the related work.
Section~\ref{s:cross-modal} describes the proposed unified
representation and the proposed descriptor. Both visual and tactile
sensing setup are described in Sec.~\ref{s:sensing}.
Section~\ref{s:results} presents diverse experiments in order 
to show the performance of the proposed cross-modal classifier.
Conclusions are reported in Sec.~\ref{s:conclusion}.

\section{RELATED WORK} \label{s:related}
While the literature of the robotics community contains a number of
contributions on both mono-modal and multi-modal object recognition
problem, the cross-modal approach has been explored mainly in the
neuroscience and psychology communities. As explained in the
introduction, a mono-modal classifier is trained with one sensing
modality and then it is queried with the same sensing modality to
recognize the observed object. Concerning the visual sensing
modality, the computer vision approach is a well-explored field and
the related literature is vast. Owing to the widespread use of
low-cost RGB-D cameras, the purely visual recognition approaches
have been accompanied by recognition algorithms based on 3D point
clouds. Specific descriptors have been proposed in the last decade
exploiting visual 3D point clouds, e.g., Persistent Feature
Histograms (PFH) \cite{rusu2008aligning}, Fast PFH (FPFH)
\cite{rusu2009fast}, Unique Signatures of Histograms (SHOT)
\cite{salti2014shot}, and Ensemble of Shape Functions (ESF)
\cite{wohlkinger2011ensemble}, and Spin Images (SI)
\cite{johnson1997representation}. A tutorial that describes and compares the most widespread descriptors is \cite{aldoma2012tutorial}.

In contrast to visual approaches, a second sensing modality widely
explored for object recognition is the tactile
perception~\cite{luo2015novel}, which can be used not only for
object classification but also for recognizing specific physical
object features~\cite{liu2012surface} such as texture or
friction~\cite{de2015integrated}. Zhang et al. \cite{zhang2016triangle} propose
a descriptor to recognize objects based on data collected by a
robotic hand equipped with tactile sensors.
In \cite{schneider2009object}, a bag-of-words approach is adopted to
recognize objects from low-resolution tactile images acquired during
the grasping with a sensorized gripper. A stochastic approach based
on the bag-of-features is proposed in~\cite{pezzementi2011tactile}
to estimate the probability distribution over object identity by
object tactile exploration.

Multi-modal perception techniques are usually adopted to improve
accuracy of the object classification algorithm by exploiting both
visual and tactile data in the training phase. The deep learning
method based on Convolutional Neural Networks (CNNs) proposed in
\cite{gao2015deep} achieves very good performance in recognizing
some material properties. The algorithm presented in
\cite{gould2008integrating} fuses visual and range data to recognize
objects, while \cite{bjorkman2013enhancing} combines visual features with
tactile glances to refine object models, obtaining more accurate
information about surfaces. In \cite{liu2017visual}, visual-tactile recognition
is carried out with 18 household objects.
Visual and tactile data are used not
only for recognition, as in \cite{ilonen2013fusing}, where 3D models
of unknown objects are reconstructed based on multi-modal data
acquired during object grasping.
In \cite{guler2014s}, both monomodal and multimodal perception are used to detect what is inside a container using robotic grasping.
The approach proposed in \cite{kroemer2011learning} exploits  visuo-tactile multimodal perception to reduce the problem of pairing, also discussed in \cite{liu2017weakly}.

Cross-modal perception has been investigated in the
neuroscience and psychology literature, e.g.,
in~\cite{cloke2015neural,schumacher2016cross}, where some studies on
animals have been carried out to demonstrate how cross-modality is
actually exploited in nature. An entire chapter
of~\cite{calvert2004handbook} is dedicate to cross-modal object
recognition. In \cite{meltzoff1979intermodal} a study is reported
concerning intermodal matching on infants, while \cite{davenport1973cross}
investigates visuo-tactile cross-modal perception in apes.

However, to the best of our knowledge, cross-modal visuo-tactile object
recognition has been investigated in the robotics community for the
first time in this work, together with our previous conference paper~\cite{falco2017cross}, which introduces the problem of
cross-modal object recognition and proposes an empirical solution.
The present paper extends the work in \cite{falco2017cross} and includes transfer learning techniques that improve significantly the accuracy of cross-modal recognition.

\section{CROSS-MODAL OBJECT RECOGNITION} \label{s:cross-modal}
This section describes the elements of our cross-modal visuo-tactile framework, i.e., unified representation, features definition, visuo-tactile transfer learning, and learning algorithm.
\subsection{Representation and Preprocessing} \label{s:unified}
The first point we address is how to represent visual and tactile data to allow an effective cross-modal perception.
RGB-D cameras allow us to represent an object $O$ as a set of points ${\cal P}=\{ \bfp_0, \bfp_1, ...,\}$, defined hereafter as point cloud of $O$. Each vector $\bfp=(p_x, p_y, p_z)$ denotes the 3D position coordinates of the point $\bfp$.
With the symbols ${\cal P}^v$ and $\bfp^v$ we indicate that the point cloud ${\cal P}$ and the point $\bfp \in {\cal P}$ is captured with visual perception. Note that the number of points of ${\cal P}$ is always different at each acquisition.
In order to derive a unified, compatible representation, {\emph{we represent tactile raw data as point clouds}}. {With raw data we mean} the contact points between the object and the sensor. {Even though representations based on tactile point clouds were used for shape reconstruction \cite{meier2011probabilistic} and creation of object bounding boxes \cite{charusta2009extraction}, this choice may appear naive for object recognition applications. In fact, modern tactile sensors can provide richer information than a point cloud, such as contact forces, textures, pressure maps, and friction coefficients. 
}
{However, as graphically shown in Fig. \ref{fig:intersection}, in order to achieve cross-modal capabilities,
a representation is required that contains information common to both visual and tactile perception.
\begin{figure}[t]
\centering
	 \includegraphics[width=\columnwidth]{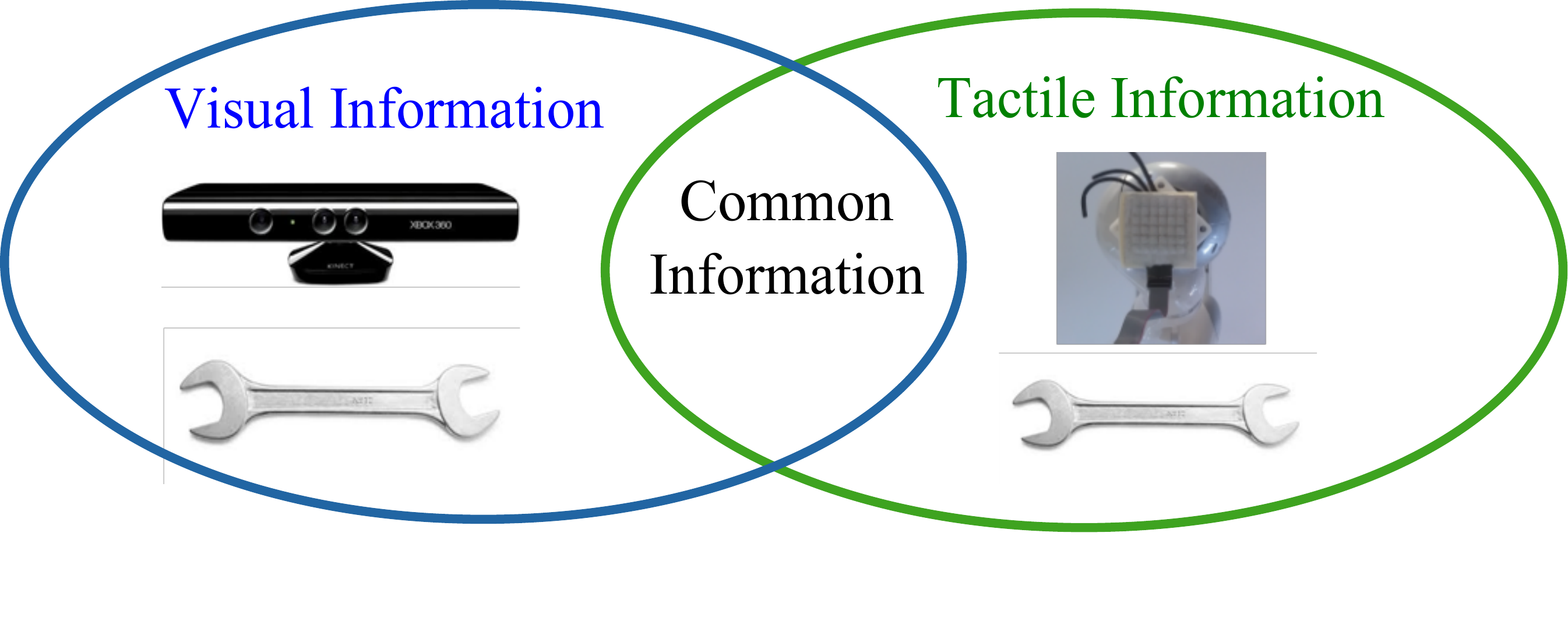}
	\caption{A representation for cross-modal perception considers common information between the modalities}
	\label{fig:intersection}
\end{figure}
{The tactile point cloud representation of the object $O$ is denoted as ${\cal P}^t=\{\bfp_0^t, \bfp_1^t, ...\}$, where the symbol $t$ denotes a point cloud acquired with a tactile perception system.
Tactile and visual point clouds present significant differences in point density, in partiality of data, and in the characteristics of noise that affects the measurements.
To derive a more effective unified representation, we equalize ${\cal P}^v$ and ${\cal P}^t$ in order to reduce the difference in point density and in partiality.
Data partiality consists in missing points in visual and tactile clouds. Even when the position and orientation of the objects are the same in both tactile and visual exploration, the tactile and visual point clouds have different missing points.
Besides partiality, visual and tactile point clouds present also different point densities.
In order to alleviate these differences, we preprocess both tactile and visual point clouds through two main steps:  \emph{equalizing partiality of the data} and \emph{uniforming point density}.
\subsubsection{Equalizing Partiality}
The method we adopt to handle data partiality is the Moving Least Squares (MLS) surface reconstruction \cite{levin2004mesh}. This step allows us to filter the measurement noise and to recreate the missing parts of the surface. The core of the MLS approach is composed by three basic steps.
We assume to have a set of points ${\cal P}$.
Given a query point $\bfp \in {\cal P}$, the first step consists in finding a plane $H$ that approximates locally the surface $S$ in a region $I$ of center $\bfp$ and radius $r$, called "search radius". The plane $H$ is computed by using Principal Component Analysis (PCA).
\begin{algorithm}[b]
	\caption{Equalization}
	\label{a:equalization}
	\begin{algorithmic}[1]
		\State function ${\cal P}={\rm equalize}(PointCloud\, {\cal P^*})$
            \State $ \bar{\cal P} = {\rm MLS}({{\cal P}^*}, u_s=0.3\,{\rm mm}, r=6\,{ \rm cm} , p_d=2)$
            \State ${\cal P}={ \rm voxelGridFilter}(l=5\, {\rm mm} )$
            \State return$(\cal P)$
	\end{algorithmic}
\end{algorithm}
{The points of the set $I$ are projected onto $H$ and upsampled with a step $u_s$ of $0.3\,$mm. With this operation, we transform the set $I$ into the set $\tilde{I}$.
The second step consists in fitting with a polynomial of order $p_d$ the height of the points projected on $H$.
We choose $p_d=2$ and $r=6\,$cm. Setting $r=6\,$cm confers rather strong filtering behavior and we can lose information in proximity of sharp edges. Typical values in monomodal visual perception are $r \in [1.5,3]\,$cm. However, in cross-modal perception, a strong filter can equalize cross-modal noise and in our case study allows achieving better performance. A more detailed and formal description of the procedure can be found in \cite{levin2004mesh}.} In this work, the parameters are chosen with a grid search approach, maximizing the recognition accuracy.
\subsubsection{Uniforming Density} The second step of the equalization procedure consists in applying a voxel grid filter \cite{voxeltutorial} to downsample and ensure a more uniform point density. We apply the voxel filtering approach implemented in Point Cloud Libraries (PCL) \cite{rusu20113d}. In this approach, the space is divided in 3D cubes (called voxels). All the points contained in each 3D box are substituted with their centroids. Following this procedure, the number of points will be equal to the number of {filled} voxels. Selecting appropriately the dimension of the voxels, the similarity of point density between tactile and visual data can be improved. In this work, we have empirically chosen cubic voxel with edge length $l=5\,$mm.

The procedure is summarized in Algorithm \ref{a:equalization}.
An example of visual and tactile point clouds before and after preprocessing is shown in Fig. \ref{FIG:point_cloud_quality}. The equalization step plays a key role in order to improve the performance (see Sec. \ref{s:results}).
\begin{figure}[t]
	 \begin{subfigure}[b]{0.24\textwidth}
		 \includegraphics[width=3cm,height=3cm,angle=180]{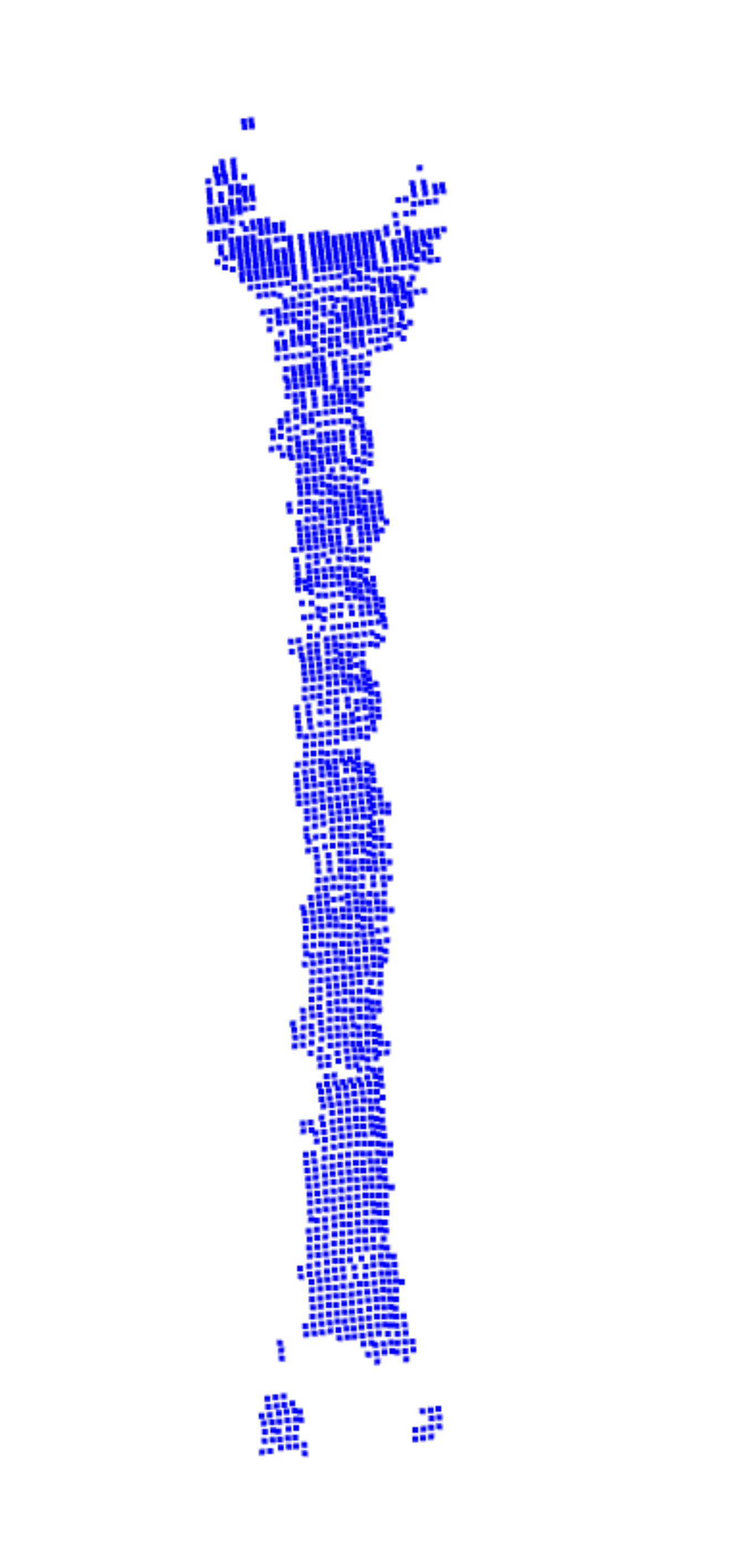}
		\caption{Visual point cloud before preprocessing}
	\end{subfigure}
	 \begin{subfigure}[b]{0.24\textwidth}
		 \includegraphics[width=3cm,height=3cm]{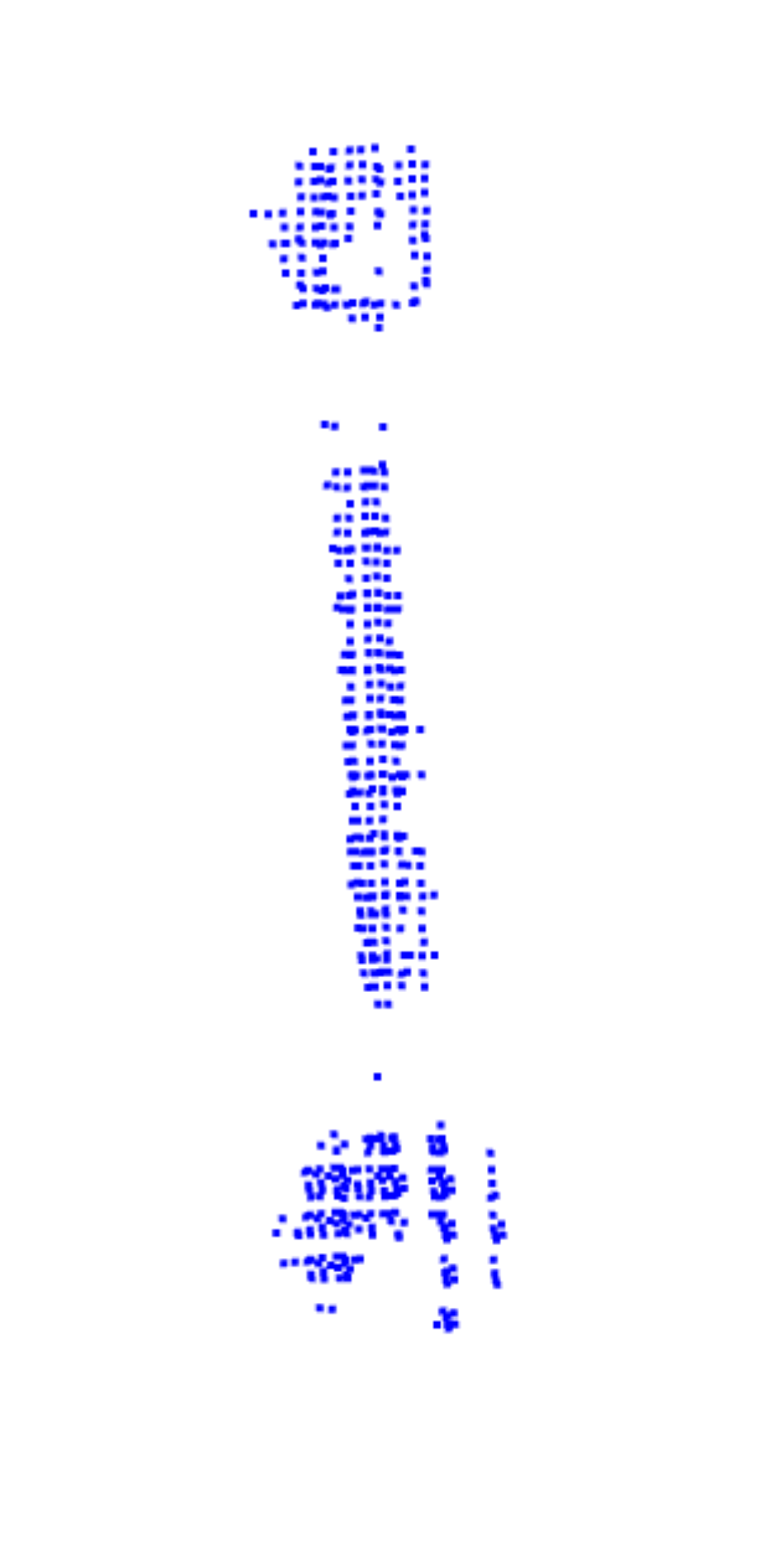}
		\caption{Tactile point cloud before preprocessing}
	\end{subfigure}
	 \begin{subfigure}[b]{0.24\textwidth}
		 \includegraphics[width=3cm,height=3cm,angle=180]{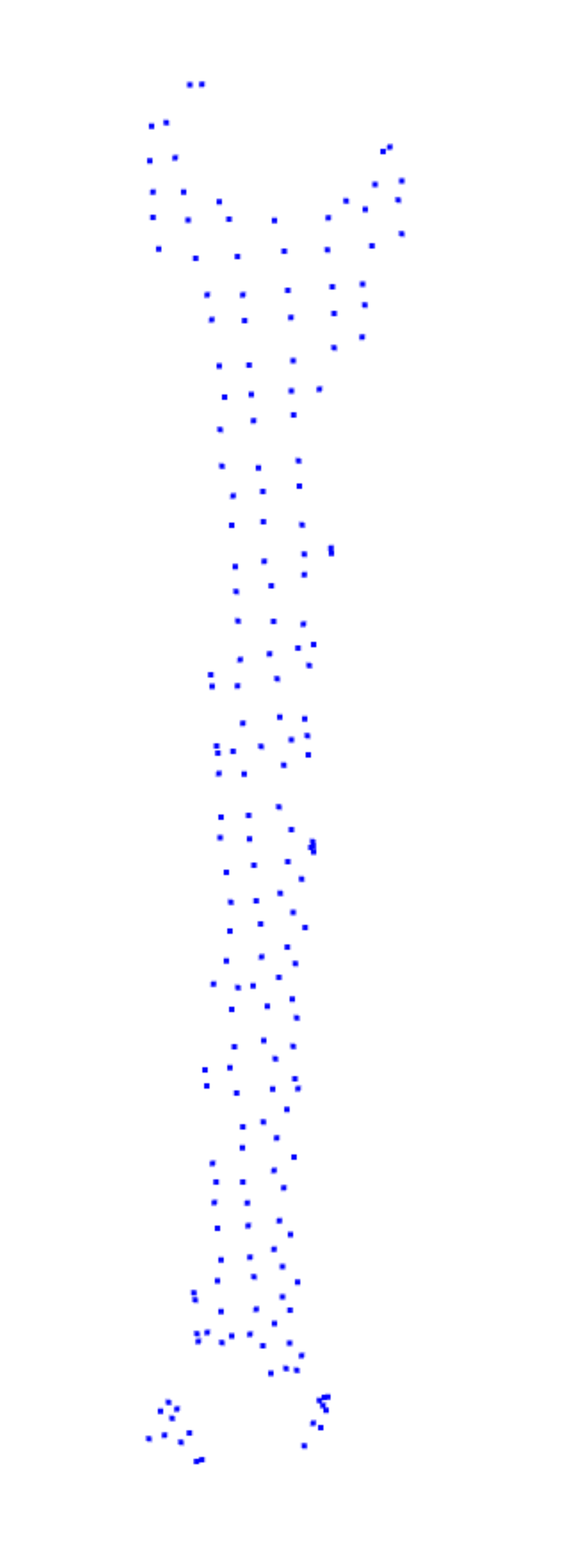}
		\caption{Visual point cloud after preprocessing}
	\end{subfigure}
	 \begin{subfigure}[b]{0.24\textwidth}
		 \includegraphics[width=3cm,height=3cm]{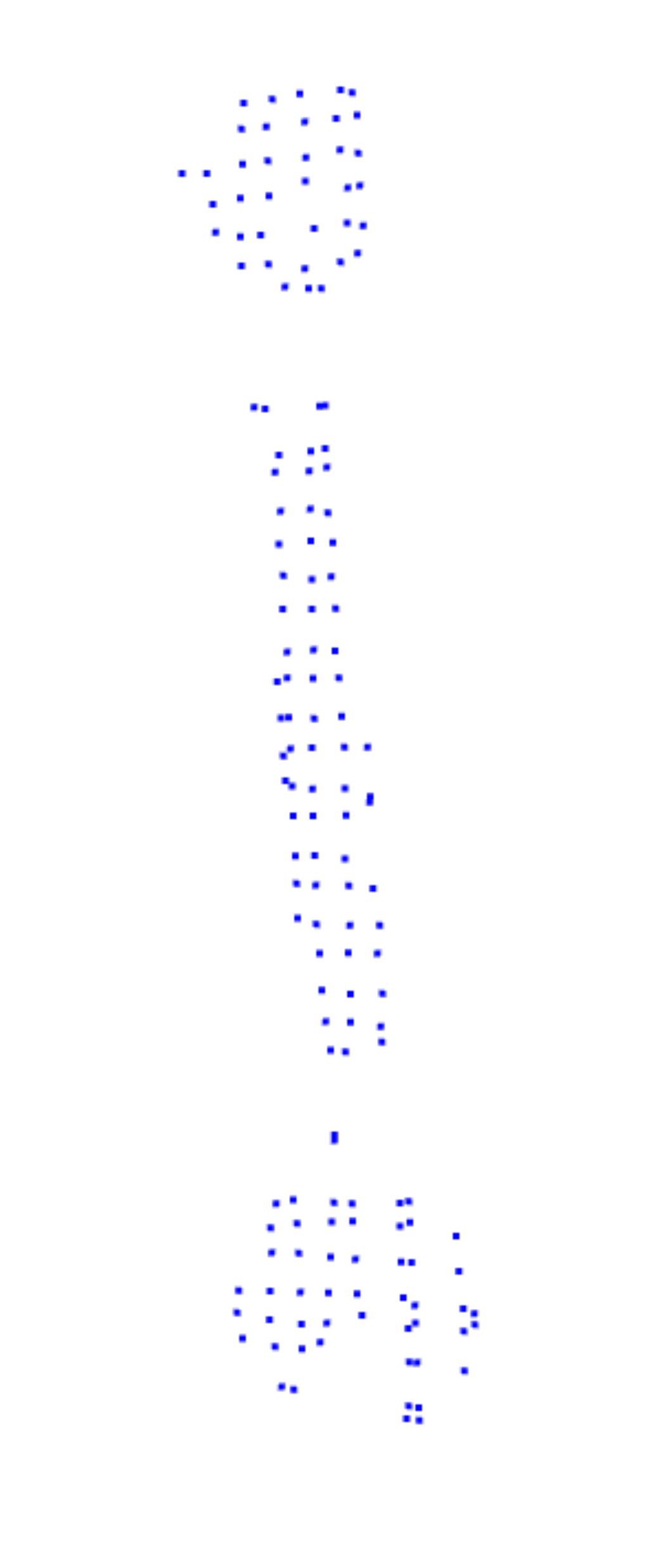}
		\caption{Tactile point cloud after preprocessing}
	\end{subfigure}
	\caption{Visual and tactile point clouds}
	\label{FIG:point_cloud_quality}
\end{figure}
\subsection{A Suitable Descriptor} \label{s:descriptor}
After defining a unified representation based on point clouds, it is important to choose a suitable feature descriptor for cross-modal recognition.
{The choice of the feature descriptors strongly depends on the chosen representation of raw data. Since our unified representation is based on point clouds, we orient our research towards 3D point cloud descriptors. From the results  reported in the computer vision literature (\cite{salti2014shot}, \cite{wohlkinger2011ensemble}, \cite{guo2016comprehensive}), SHOT and ESF seem promising candidates for our problem.}
{However, our experimental data shows that even after the preprocessing step, the differences in noise, resolution and partiality of the data found in the training and test set cannot be equalized perfectly. Therefore, there is a need of a new, robust descriptor suitable for cross-modality.}
Following a strategy commonly adopted in communication engineering, we propose to increase the redundancy of the information associated to the descriptors. A way to increase the redundancy is finding a smart combination of different descriptors. {In our case we expect benefit by combining SHOT and ESF, since they encode information with two different approaches, i.e. normal-based and normal-free respectively. In particular, SHOT encodes point clouds with histograms of normal vectors \cite{salti2014shot}, while ESF encodes information based on shape functions \cite{wohlkinger2011ensemble}. Also, they both present the best performance according to our comparisons (Table \ref{t:result_with_preprocessing}) and do well according to works in the literature, e.g., \cite{guo2016comprehensive,wohlkinger2011ensemble}.} 
The ESF descriptor is an ensemble of 10 concatenated histograms of shape functions consisting of angle, point distance, and area functions. Each histogram has 64 bin, for a total of 640 elements \cite{wohlkinger2011ensemble}.
The SHOT computes a local reference frame $\Sigma_c$ using the eigenvalue decomposition around an input point $\bfc$, in our case $\bfc$ is the centroid of the point cloud $\cal P$. Given the frame $\Sigma_c$, a sphere $S_c$ of center $\bfc$ and radius $r_c$  is defined. $S_c$ is then split into 32 divisions and for each division a 11-bin histogram is computed. Each histogram contains angles that describe the directions of the normal vectors to each point $\bfp \in \cal P$ in the frame $\Sigma_c$. The descriptor concatenates the histograms into the final signature, obtaining a vector of 352 elements. We compute a single SHOT feature for each object and use it as a global feature.

Let $\bfd_{SHOT}$ be the SHOT descriptor associated to the point cloud $\cal P$ and $\bfd_{ESF}$ the ESF descriptor associated to the same point cloud. Both descriptors are {column} vectors.
{
The first possible way to improve the performance is to simply concatenate $\bfd_{SHOT}$ and $\bfd_{ESF}$ so that:
\begin{equation} \label{eq:naive}
    \bfd_c=[\bfd_{SHOT}^T \quad \bfd_{ESF}^T]^T,
\end{equation}
where $\bfd_c$ is the concatenated descriptor. Even if the concatenated descriptor $\bfd_c$ contains more information than $\bfd_{ESF}$, the improvement in accuracy was not significant.
This can happen because the dimension of $\bfd_c$ is much higher than both SHOT and ESF. As a consequence, the classification problem is affected by the curse of dimensionality.
Moreover, the high increase in dimension can be a limitation also in terms of training time and classification time, especially when scaling to very large databases. The SHOT descriptor associated to a point cloud is a vector of 352 elements, while an ESF descriptor is a vector of 640 elements.
Therefore, we decided to exploit a data compression method. {
To compress the descriptor $\bfd_c$, we organize the vectors $\bfd_{ESF}$ and $\bfd_{SHOT}$ in the matrix:
\begin{equation}\label{eq:D}
\hat{\bfD}=[\bfd_{ESF} \quad  \tilde{\bfd}_{SHOT}],
\end{equation}
where $\tilde{\bfd}_{SHOT}=[{\bfd}_{SHOT}^{\rm T} \quad \mathbf{\bar{0}}]^{\rm T}$ and $\mathbf{\bar{0}}$ is a 0-vector of dimension $1 \times (640-352)$.} We want to compress the information carried from the matrix $\hat{\bfD} \in \mathbb{R}^{640 \times 2}$ into a vector $\bfd_r \in \mathbb{R}^{640}$.}
{We leverage the data compression capability of Singular Value Decomposition (SVD) \cite{kalman1996singularly}. First, we center the matrix $\hat{\bfD}$, so that all columns are zero-mean. Let $\bfD$ be such a mean-centered matrix. 
The SVD of the matrix $\bfD$ is}
\begin{equation}\label{eq:svd_def}
\bfD=\bfU \bf\Sigma \bfV^T,
\end{equation}
{where $\bfU=[\bfu_1, \bfu_2, .... \bfu_{640}] \in \mathbb{R}^{640 \times 640}$, $\bfV=[\bfv_1, \bfv_2] \in \mathbb{R}^{2 \times 2}$, and $\bfSigma \in \mathbb{R}^{640 \times 2}$ is the matrix that contains the singular values $\sigma_1$ and $\sigma_2$, $\bfU$ and $\bfV$ contain the left and right singular vectors, respectively.
We choose the compressed SVD descriptor $\bfd_r$ as}
\begin{equation}\label{eq:svd}
\bfd_r=\sigma_1\bfu_1.
\end{equation}
{Since we make zero-mean the columns of the matrix $\bfD$, the singular value decomposition is equivalent to a decomposition based on principal component analysis.}
This descriptor has dimension $640$ and is a linear combination of the columns of the matrix $\bfD$.
The best rank-$1$ approximation of the matrix $\bfD$ is given by the matrix $\bfD_1=\sigma_1 \bfu_1 \bfv_1^{\rm T}$. As a consequence, $\bfd_r \bfv_1^{\rm T}$ is the 1-rank matrix that minimizes the norm $\|\bfD-\bfD_1\|$. 

Using $\bfd_r$ as a descriptor we obtain significantly better performance than using the $\bfd_c$ \cite{falco2017cross}. The descriptor $\bfd_r$, in fact, carries more information than both {ESF or SHOT}, but is less affected by the curse of dimensionality than the concatenated descriptor $\bfd_c$.
We call the descriptor derived in Eq. (\ref{eq:svd}) Cross-modal point cLoUd dEscriptor (CLUE). The CLUE descriptor consists in the basic ESF enriched with the information carried by SHOT. CLUE performs better than both ESF and SHOT in the cross-modal case, while it performs very similarly in the monomodal and multimodal cases. Hence, we guess that the robustness of combining a normal-based descriptor (SHOT) and a normal-free descriptor (ESF) is beneficial for cross-modal applications in particular.

\subsection{Transfer Learning} \label{sec:TL}
Exploiting the equalization step and the CLUE descriptor, we obtain a significant improvement in the accuracy \cite{falco2017cross}. In order to further improve the accuracy, we  propose to adopt techniques from transfer learning. Transfer learning approaches have been adopted effectively in computer vision and textual document recognition \cite{pan2010survey}.

In Transfer Learning, and in particular domain adaptation, we define a source domain $\mathcal{D}_S$ and a target domain $\mathcal{D}_T$. A generic domain $\mathcal{D}$ is constituted by the couples $(\bfx, p(\bfX))$, where $\bfx \in X$ is the features vector, $p(\bfX)$ is the probability distribution of the feature space. In this work, the vector $\bfx$ is constituted {by the elements of} the descriptor, e.g., CLUE shown in Sec. \ref{s:descriptor}.
Beside a domain, we define also a learning task $\mathcal{T}=\{\bfY, f(\cdot)\}$, where $\bfY$ is a set of labels and $f: \bfx \in \mathcal{X} \rightarrow y \in \mathcal{Y}$ is the function which associates feature vectors to classes (or labels).
The source dataset $ D_S=\{\bfx_{S_1}, ..., \bfx_{S_N}\}$  is constituted by the source feature vectors. Moreover, we denote with $\tilde{D}_S=\{(\bfx_{S_1}, y_{S_1}), ..., (\bfx_{S_N}, y_{S_N})\}$ the source dataset with labels.
The target dataset $D_T= \{(\bfx_{T_1}), ..., (\bfx_{T_N})$ is constituted only by feature vectors but typically not by labels.
Moreover, we can define a source task $\mathcal{Y_S}=\{Y_S, f_S(.)\}$ and a target task $\mathcal{Y_T}$.
Given a source domain $\mathcal{D}_S$, a source learning task $\mathcal{T}_S$, a target domain $\mathcal{D}_T$, and a target learning task $\mathcal{T}_T$, transfer learning aims to improve the learning of $\mathcal{T}_T$ by using the knowledge of $\mathcal{D}_S$ and $\mathcal{T}_S$. When $\mathcal{T}_T = \mathcal{T}_S$, as in our case, we have a domain adaptation problem.

We focus on using the visual data as source domain and the tactile data as target domain. 
The principal reason is that it is easier, in general, to collect several images and it is more demanding to collect many tactile examples. Collecting tactile data requires physical interaction with the external environments and, on the long run, devices such as sensors can be damaged or a significant amount of time or energy can be required for haptic exploration.
In simpler words, the transfer learning problem can be formulated this way: given the knowledge of the labeled set $\tilde{D}_S$, can we estimate labels for the set $D_T$?

In the literature, transfer learning approaches have been used in text classification and in image recognition to transfer knowledge from a data base to another, exploiting the same perception modality.
In our case, we investigate if such techniques can be adopted to transfer knowledge across different perception modalities.
To this aim, we apply to our problem two classes of approaches. The first class is based on dimensionality reduction, which includes Transfer Component Analysis (TCA) \cite{pan2011domain} and Subspace Alignment (SA) \cite{fernando2013unsupervised}.
The second class is based on the geodesic flow associated to different subspaces. This class includes Geodesic Flow Kernel (GFK) \cite{gong2012geodesic}.
We briefly describe here 
the GFK approach.
For details on TCA and SA, please refer to \cite{pan2011domain} and \cite{fernando2013unsupervised}, respectively.

Existing approaches such as TCA and SA focus on learning feature representations that are invariant across domains. The basic idea of GFK is to integrate an infinite number of subspaces that characterize changes in geometric and statistical properties from the source domain to the target domain.
For our application in cross-modal object recognition, GFK shows the best performance, and combined with equalization of partiality and resolution, achieves similar performance of monomodal recognition and slightly outperforms visuo-tactile multimodal recognition.
The method we used in this work is based on  \cite{gong2012geodesic}.
In order to apply GFK, we introduce the concept of \emph{Grassmann Manifold}.
Given a vector space $\mathcal{D}$ of dimension $D$, a Grassmannian $G(d, D)$ is a manifold which includes all $d$-dimensional linear subspaces of the vector space $\mathcal{D}$. In our case, the source domain is denoted $\mathcal{D}_S$ and the target domain is denoted as $\mathcal{D}_T$. Both the spaces have dimension $D$.
The first step of GFK transfer learning consists in reducing the dimensionality of both source and target domains    with a linear operator. A typical option is principal component analysis:
\begin{eqnarray}
  X_S &=& {\rm PCA}(D_S,d) \\
  X_T &=& {\rm PCA}(D_T,d),
\end{eqnarray}
where, similarly as in SA, $d$ is a parameter of the algorithm. The matrix $X_S \in \mathbb{R}^{D \times d}$ represents the basis for the linear subspace of $\mathcal{D}_S$ obtained through PCA. Similarly, the matrix $X_T \in \mathbb{R}^{D \times d}$ represents the PCA basis for the linear subspace of $\mathcal{D}_T$.
Since $X_S$ and $X_T$ can be seen as $d$-dimensional subspaces of a $D$-dimensional space, they are points of a Grassmann Manifold $G(d, D)$.  Let $R_S$ and $R_T$ be the orthogonal complements to $X_S$ and $X_T$, respectively.
The geodesic flow is represented with the parametric function
\begin{equation}\label{eq:geodesic_flow}
    \Phi:t \in [0,1] \rightarrow \Phi(t) \in G(d,D),
\end{equation}
where
\begin{eqnarray} \label{eq:geodesic_flow_conditions}
  \Phi(0) &=& X_S \\
  \Phi(1) &=& X_T.
\end{eqnarray}
The parametric function $\Phi(t)$, in brief, maps the values of the parameter $t$ to all the subspaces that connect the source domain and the target domain.
In particular, as shown in \cite{gong2012geodesic}, the function $\Phi$ has the form:
\begin{equation}\label{eq:Phi}
    \Phi(t)=P_SU_1\Gamma(t)-R_S U_2 \Sigma(t),
\end{equation}
where $\Gamma(t)={\rm diag}\{\sin(t\theta_1), \sin(t\theta_2), ..., \sin(t\theta_d)\}$, $\Sigma(t)={\rm diag}\{\cos(t\theta_1), \cos(t\theta_2), ..., \cos(t\theta_d)\}$, and $\{\theta_1, \theta_2 \, ...\, \theta_d\}$ are the principal angles between the source and target domains. It holds $\theta_1<\theta_2 < \, ...\, \theta_d\ < \pi/2$.
The matrices $U_1 \in \mathbb{R}^{d \times d}$ and $U_2 \in \mathbb{R}^{(D-d) \times d}$ are computed by the following singular value decompositions
\begin{eqnarray}
  X_S^{{\rm T}} X_T &=& U_1 \Gamma V^{{\rm T}} \\
  R_S^{{\rm T}}P_T &=& -U_2 \Sigma V^{{{\rm T}}}.
\end{eqnarray}

After defining the function $\Phi$, the GFK approach uses the infinite-dimensional feature vector $z^{\infty}$ defined as
\begin{equation}\label{eq:infinity}
    \bfz^{\infty}=[\Phi(0),..., \Phi(t), ..., \Phi(1) ]^{ {\rm T} }\bfx,
\end{equation}
where $t \in [0,1]$, $\bfx$ is the feature vector as derived in Sec. \ref{s:descriptor}, e.g. CLUE.
Since the vector $\bfz^{\infty}$ has infinite dimension, it is not usable directly by a digital computer.
However, exploiting the kernel trick \cite{bishop2006pattern}, we can define a distance between two infinite-dimensional vectors $\bf z^{\infty}_i$ and $\bfz^{\infty}_j$ using the scalar product
\begin{equation}\label{eq:kernel_trick}
    \langle \bfz^{\infty}_i, \bfz^{\infty}_j\rangle= \int_{0}^1(\Phi(t)^{\rm T}\bfx_i)^{\rm T}( \Phi(t)^{\rm T}\bfx_j)dt.
\end{equation}
It is possible to show that the integral in Eq. (\ref{eq:kernel_trick}) is equal to the following quadratic form:
\begin{equation}\label{eq:quadratic_form}
  \int_{0}^1(\Phi(t)^{\rm T}\bfx_i)^{\rm T}( \Phi(t)^{\rm T}\bfx_j)dt=\bfx_i^{\rm T}\bfG\bfx_j,
\end{equation}
where the matrix $\bfG \in \mathbb{R}^{D \times D}$ is such that
\begin{equation}\label{eq:Gmatrix}
G= \begin{bmatrix} \bfX_S \bfU_1 & \bfR_S\bfU_2
   \end{bmatrix}
  \begin{bmatrix}
    \bfLambda_1 & \bfLambda_2\\
    \bfLambda_2 & \bfLambda_3
  \end{bmatrix}
  \begin{bmatrix}
    \bfU_1^{ {\rm T}} \bfX_S^{ {\rm T}} \\
     \bfU_2^{ {\rm T}} X_S^{ {\rm T}},
  \end{bmatrix}
\end{equation}
with
\begin{equation}\label{eq:Lambda}
\bfLambda_i={\rm diag}\{\lambda_{i_1}, \lambda_{i_2}, ..., \lambda_{i_d}\}, \quad i=1,2,3
\end{equation}
and
\begin{eqnarray}
  \lambda_{1_j} &=& 1+ \frac{\sin(2\theta_j)}{2 \theta_j} \\
  \lambda_{2_j} &=& \frac{\cos(2\theta_j)-1}{2 \theta_j} \\
  \lambda_{3_j} &=& 1- \frac{\sin(2\theta_j)}{2 \theta_j}, \quad j=1,2,...,d.
\end{eqnarray}

It is interesting to note that the GFK approach compute the similarity between two feature descriptors by leveraging the matrix $\bfG$. Such matrix depends on both the source domain and the target domain. When $\bfG=\bfI$, the similarity becomes the classical scalar product in the Euclidean space.
The sole parameter of the algorithm is the subspace dimension $d$. Detailed guidelines on how to tune this parameter are reported in \cite{gong2012geodesic}.
{}It is important to remark that the GFK transfer learning algorithm exploits both data from the source domain and data for the target domain. 
However, none of such data is labelled. Data from the target domain, in fact, are used only in the adaptation phase in unsupervised fashion and not in the training phase. }
\begin{algorithm}[tb]
	\caption{CMR Training}
	\label{a:CMR_training}
	\begin{algorithmic}[1]
		\State function ${\cal M}^v = {\rm training}({\cal P}_S, {\rm Labels}\, Y_S)$
        \State $\bar{\cal P}_S={\rm equalize}({\cal P}_S)$
        \State $D_S= {\rm computeCLUE}(\bar{\cal P}_S)$
        \State ${\rm Model}\,{\cal M}^v= {\rm train}(D_S, Y_S)$
        \State return$({\cal M}^v)$
	\end{algorithmic}
\end{algorithm}

\begin{algorithm}[tb]
	\caption{TL-CMR Training}
	\label{a:TL-CMR_training}
	\begin{algorithmic}[1]
		\State function $ {\cal M}^v={\rm training}({\cal P}_S, {\cal P}_T, {\rm Labels}\, Y_S)$
        \State $\bar{\cal P}_S= {\rm equalize}({\cal P}_S)$
        \State $\bar{\cal P}_T= {\rm equalize}({\cal P}_T)$
        \State $D_S = {\rm computeCLUE}(\bar{\cal P}_S)$
        \State $D_T={\rm computeCLUE}(\bar{\cal P}_T)$
        \State $\bfG={\rm GFK}(D_S,D_T,d)$
        \State $ {\rm Model}\,{\cal M}^v= {\rm train}(D_S, Y_S, \bfG)$
        \State return$({\cal M}^v)$
	\end{algorithmic}
\end{algorithm}

\subsection{Classification Algorithm} \label{s:learning_algorithm}
We compare $k$-Nearest Neighbor ($k$-NN), with different values of $k$ and radial basis function kernel Support Vector Machines (RBF-SVM), and linear SVM.
Both $k$-NN and SVM are simple and widely-used algorithms for classification problems.
We apply such learning algorithms to several state-of-the-art visual descriptors and with the one proposed in this work.

In more detail, to deal with the cross-modal recognition problem, we perform two steps. The first step consists in building a model ${\cal M}^v$, which embeds a-priori knowledge derived from visual perception. The second step is to exploit a-priori knowledge embedded in ${\cal M}^v$ with data from a different sensing modality.
In this work, we build the model by following and comparing two procedures, i.e., Cross Modal Recognition (CMR) pipeline and Transfer Learning-based CMR (TL-CMR). For CMR, we need only labeled source-domain data to build the model, while for TL-CMR we need labeled source-domain data and a set of unlabeled target-domain data.
\begin{algorithm}[b]
	\caption{Cross-modal Recognition}
	\label{a:encoding}
	\begin{algorithmic}[1]
		\State function $l={\rm recognize}(PointCloud\, {\cal P}^t_o, {\rm Model }\,{\cal M}^v$)
            \State $\bar{\cal P}^t_o=$ equalize(${\cal P}^t_o$)
            \State $\bfd_{SHOT}= {\rm computeSHOT}(\bar{\cal P}^t_o)$
            \State $\bfd_{ESF}= {\rm computeESF}(\bar{\cal P}^t_o)$
            \State $\hat{\bfD}=[\bfd_{SHOT} \quad \bfd_{ESF}]$
            \State $\bfD=\rm{center}(\hat{\bfD})$
            \State $[\bfU,\bf\Sigma,\bfV]=\rm{svd}(\bfD)$
            \State $\bfd_{CLUE}=\bfU(:,1)\bf\Sigma (1,1)$
            \State $l={\rm classify}(\bfd_{CLUE}, {\cal M}^v)$
            \State return$(l)$
	\end{algorithmic}
\end{algorithm}
\subsubsection{Building the model using CMR}
We use visual point clouds of $15$ objects and for each object we collect $40$ examples. Each example $i$ consists in a point cloud ${\cal P}_i$. For each point cloud ${\cal P}_i$, we apply the equalization procedure described in Sec. \ref{s:unified} and compute the CLUE descriptor $\bfd_i$, which is the representation of the point cloud in the feature space. Let $D_S$ be the set of the CLUE descriptors associated to all the examples, i.e., $ D_S = ( \bfd_1, \bfd_2, ... \bfd_n )$, with $n=600$ in our case. Also, let $Y_S$ be the set of all the labels associated the elements of $D_S$. The labeled source dataset is denoted as $\tilde{D}_S=(D_S,Y_S)$. We derive the model ${\cal M}^v$ using the set $\tilde{D}_S$. When using $k$-NN, the model ${\cal M}^v$ simply consists of the elements of $D_S$. In case of SVM and other methods that require an explicit training step, we use $D_S$ as training set and the model ${\cal M}^v$ is the trained classifier. The procedure to build a CMR model is described in Algorithm \ref{a:CMR_training}. The sensing system and the procedure to collect visual data is described in Sec. \ref{s:visual_sensing}.
We call this process Cross Modal Recognition (CMR). In this case, we use only labeled source training examples to build our model.

 \subsubsection{Building the model using TL-CMR}
 Besides the CMR pipeline, we exploit the transfer learning techniques described in Sec. \ref{sec:TL} to improve the performance.
 In order to build a model with the SA and GFK approaches we need not only the labeled source dataset $\tilde{D}_S$, but also unlabeled examples of the target domain $D_T$.
 Hence, when applying TL-CMR we assume  that the robot has haptically explored the objects of the target dataset but it is not provided with labels.
 The procedure of building the model with TL-CMR is described in Algorithm \ref{a:TL-CMR_training}.
 The TL-CMR procedure consists in building the source labeled dataset $\tilde{D}_S$ by using equalization procedure and descriptor computation on the acquired point clouds. Then, the target unlabeled dataset $D_T$ is built applying equalization and descriptor computation to the target unlabeled point clouds.
 The next step is to apply  the transfer learning algorithm to the unlabeled source data $D_S$ and the unlabeled target domain data $D_T$. The output of the transfer learning algorithm is a domain adaptation factor. In training the classifier, the adaptation factor is taken into account to reduce the differences between target and source domain. In the case of GKF, the adaptation factor is the matrix $\bfG$, which replaces the scalar product with a quadratic form to compute the similarity between two feature vectors \cite{gong2012geodesic}. In the case of SA, we have a matrix $\bfM$ that aims at aligning source and target domains in a reduced subspace, as described in Sec. \ref{sec:TL}. 
 {In order to apply the overall TL-CMR pipeline we need labeled source-domain data and unlabeled target-domain data. Target-domain data are in fact exploited only for the adaptation step and not for model training. If collecting a few unlabeled examples of the target domain is not possible for a specific case study, the simpler CMR pipeline should be used.}
\subsubsection{Exploiting the model for cross-modal recognition}
We exploit the knowledge accumulated with visual perception in order to interpret tactile data at execution time. To test the performance of cross-modal recognition, we classify the outcome of $5$ tactile explorations per object. The tactile sensing system and the exploration procedure are described in Sec. \ref{s:tactile_sensing}. After the acquisition of the tactile point cloud,  we derive the descriptor $\bfd$ with the procedure described in Sec. \ref{s:descriptor}. To recognize the object through visual a-priori knowledge, we provide $\bfd$ as an input to the classifier which embeds the model ${\cal M}^v$. The output of such a classifier is the estimated class of the object.

The entire process of visuo-tactile recognition that adopts the CLUE descriptor is summarized in Algorithm \ref{a:encoding}. The inputs of the algorithm are the model ${\cal M}^v$, derived by visual data a-priori known and the point cloud observed by tactile sensors ${\cal P}^t_O$ at execution time. The output is the label $l$ of the explored object $O$.

\begin{figure}[b]
\,
\begin{subfigure}[b]{0.4\columnwidth}
\centering
\includegraphics[height=4cm,angle=-90]{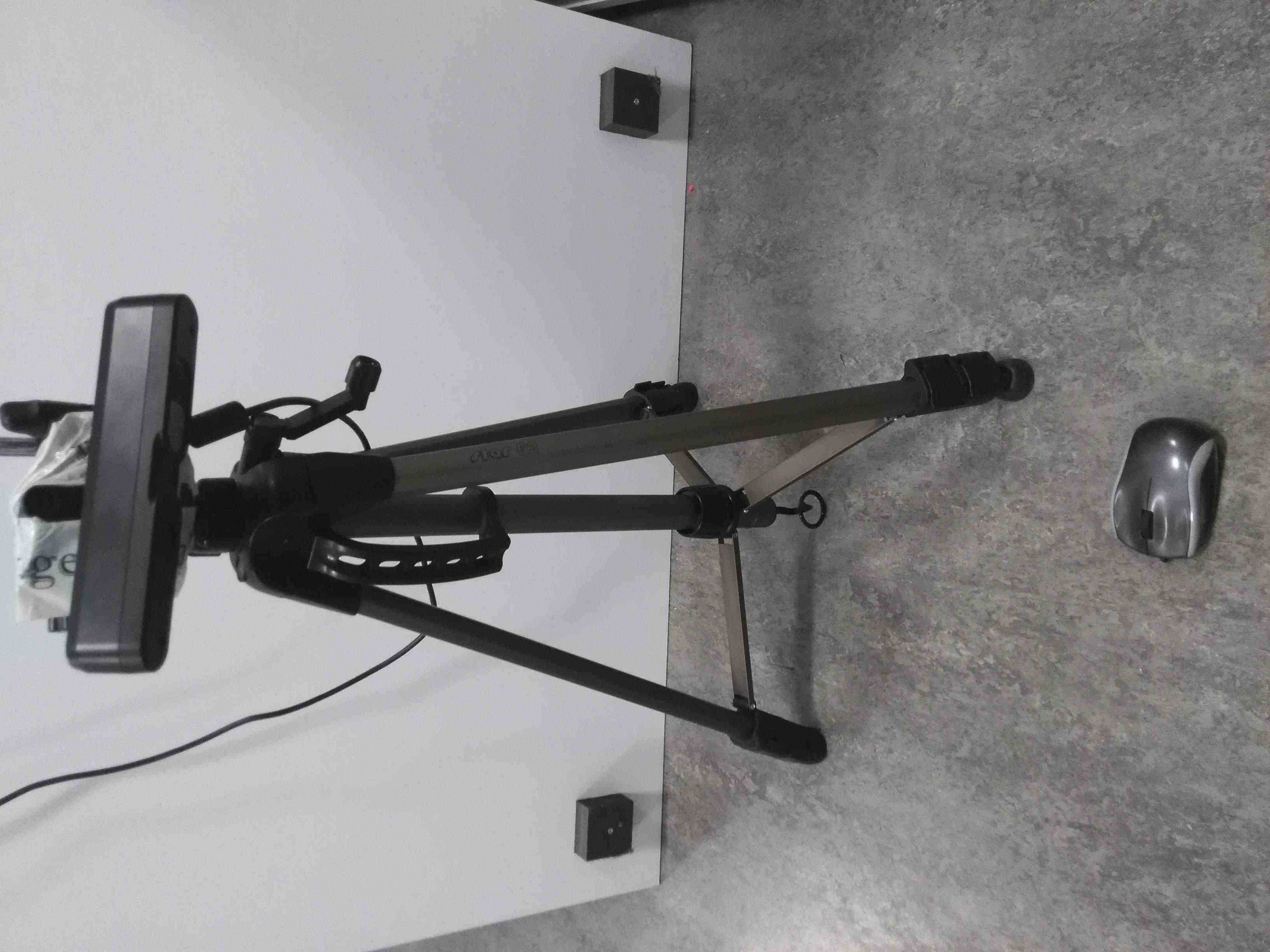} 
\caption{Visual Setup}
\label{t:visual_sensing}
\end{subfigure}
\qquad \quad
\begin{subfigure}[b]{0.45\columnwidth}
\begin{minipage}[t]{0.82\linewidth}
	\begin{tikzpicture}
	\node[anchor=south west, inner sep=0] at (0,0){\includegraphics[width=\textwidth]{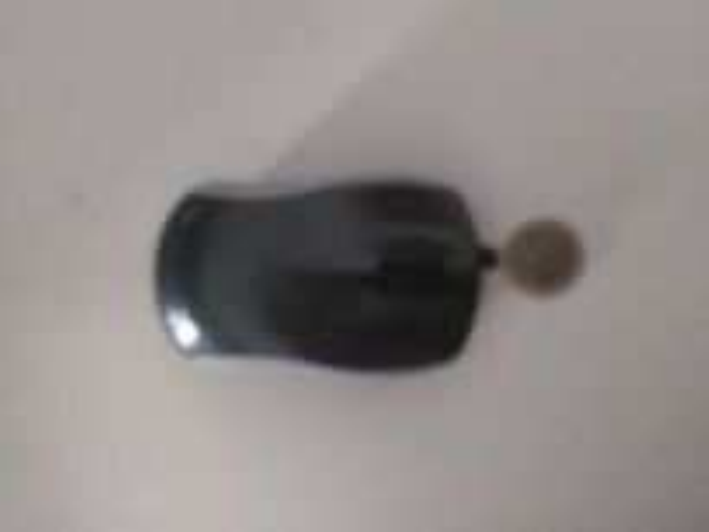}};
    \begin{scope}
    \draw (current bounding box.south west)
    grid [step=0.5cm] (current bounding box.north east);
    \end{scope}
	\end{tikzpicture}
	\caption{Tactile exploration grid}
	\label{FIG:exploration_path}
	\end{minipage}
\\
\begin{minipage}[b]{0.82\linewidth}
	\begin{tikzpicture}
	\node[anchor=south west, inner sep=0] at (0,0){\includegraphics[width=\textwidth,angle=180]{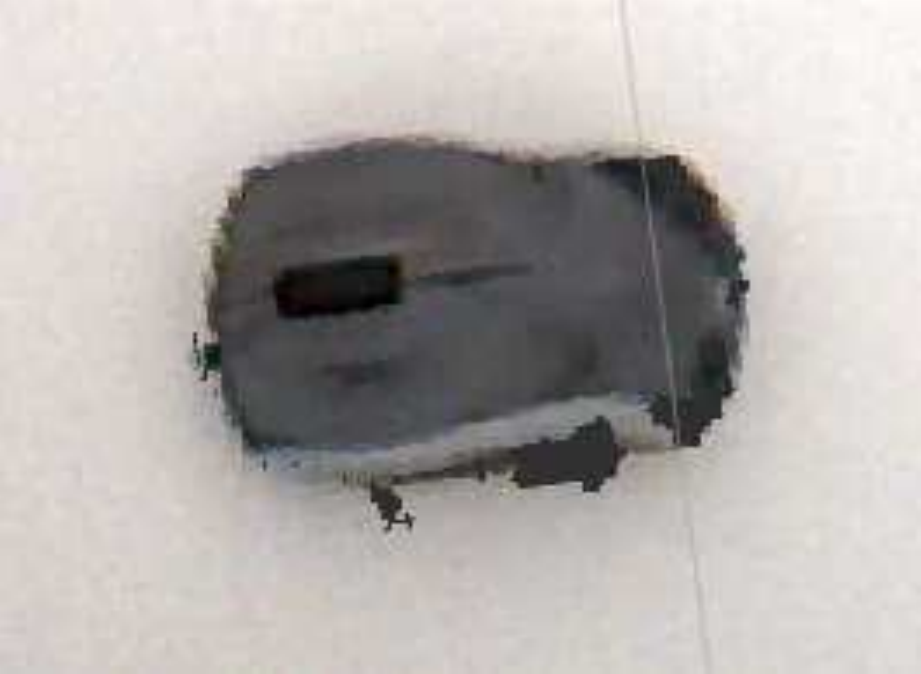}};
	\end{tikzpicture}
	\caption{Visual Point Cloud}
	\label{Fig:visual_point_cloud}
	\end{minipage}
\end{subfigure}
\caption{Visual Sensing System}
\label{FIG:visual_setup}	
\end{figure}
\section{SENSING SYSTEM} \label{s:sensing}
An experimental setup has been prepared in order to test the
effectiveness of the cross-modal object recognition approach
proposed in Sec.~\ref{s:cross-modal}. The system is constituted by a
robotic arm (KUKA LWR-IV) equipped with a tactile sensor and an
external visual perception system.

\subsection{Visual Perception} \label{s:visual_sensing}
Figure ~\ref{FIG:visual_setup} shows the visual perception system,
constituted by an {Asus Xtion Pro Live} RGB-D camera, which has been used to collect
the visual point clouds. {All relative positions between camera and
objects are as shown
Fig.~\ref{FIG:visual_setup}.} For an object $O$ the collected point
cloud is separated from the rest of the scene by using an ECE
(Euclidean Cluster Extraction) algorithm, available from the PCL
libraries. This algorithm removes the planes from the scene and
clusters the remaining points by using a kd-tree approach.
{Each object has been placed in two different poses during the acquisitions to add more variability to the data. The descriptors used in our approach, though, are invariant to position and orientation of the objects.}

\subsection{Tactile Perception} \label{s:tactile_sensing}
\subsubsection{Tactile Sensing Setup}
In the experimental setup, the tactile skin developed within the
SAPHARI Project~\cite{cirillo2016conformable} has been fixed on the
end effector of the KUKA robotic arm, as shown in
Figs.~\ref{fig:robot},~\ref{fig:eeskin}.

The distributed tactile sensor, originally presented
in~\cite{cirillo2014artificial}, consists of a PCB (Printed Circuit
Board) constituted by couples (emitter/detector) of optoelectronic
devices, used to detect the local deformations of the deformable
layer covering the optoelectronic layer. These deformations are
related to the external contact forces applied to the sensor.

The used sensor is constituted by an interconnection of a number of
identical sensing modules, each capable to estimate the three
components of the contact force applied to it. In particular, each
sensing module is constituted by four optoelectronic couples
organized in a $2\times2$ matrix. The whole sensor consists of a
$6\times6$ grid of sensing modules with a total size equal to
$5\times5\,$cm$^2$. Each sensing module, shown in
Figs.~\ref{fig:skinforce},~\ref{fig:skintactile}, has a unique
spatial representation in the robotic base frame and provides the
estimated three force components.

The $i$th contact point $\bfp_i$ is selected, when the contact force
intensity $\|\bfF_i\|$ estimated by the $i$th module is larger than
a threshold $\beta$. For the experimental results, reported in this
paper, the threshold value has been empirically chosen as
$\beta=0.8\,$N. Then, for each object, the tactile readings are
represented as three-dimensional point clouds, as described in
Sec.~\ref{s:cross-modal}.

  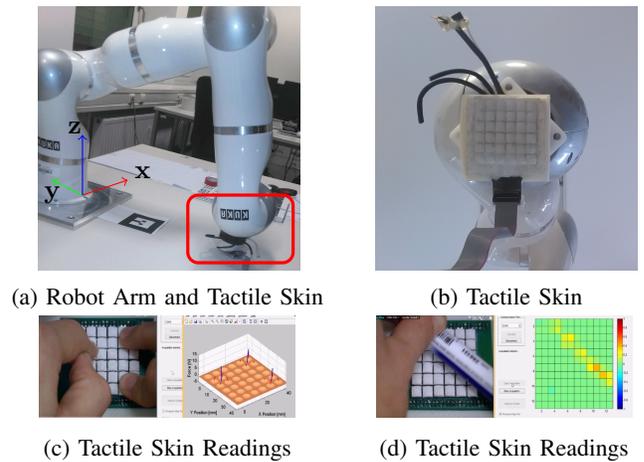
\begin{figure}[tb]
     \begin{subfigure}[t]{0.24\textwidth}
    \centering
        \begin{tikzpicture}
        \node[anchor=south west, inner sep = 0] at (0,0) {\includegraphics[width=0.8\textwidth,height=3.5cm]{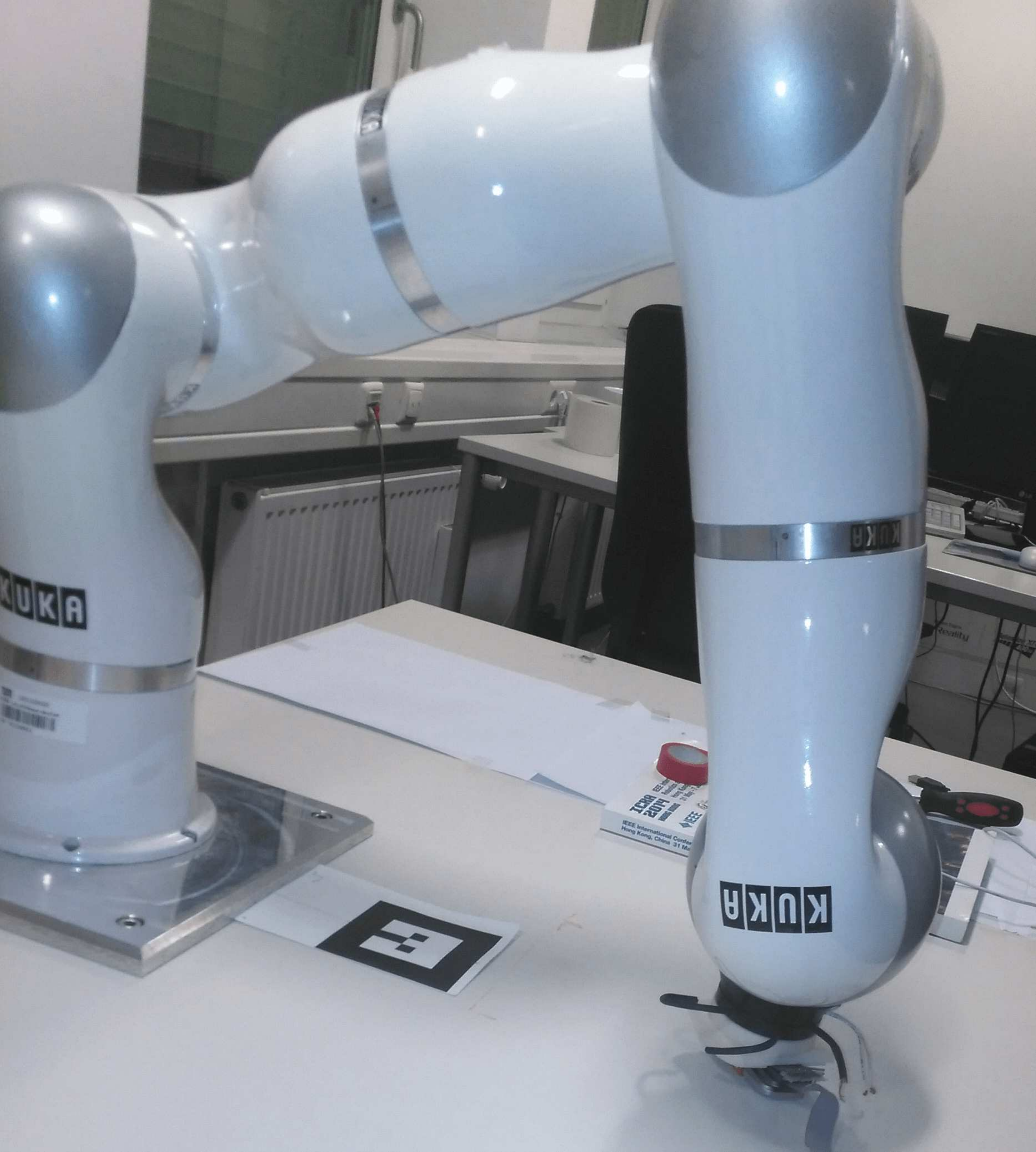}};
        \draw[->,green] (0.6,1.0)--(0.2,1.2);
        \draw[->,red] (0.6,1.0)--(1.2,1.2);
        \draw[->,blue] (0.6,1.0)--(0.6,1.8);
        \node at (0.2,1) {$\mathbf{y}$};
        \node at (1.4,1.3) {$\mathbf{x}$};
        \node at (0.5,1.9) {$\mathbf{z}$};
        \draw[red,rounded corners,very thick] (2.0,0.1) rectangle (3.4,1);
        \end{tikzpicture}
        \caption{Robot Arm and Tactile Skin}
        \label{fig:robot}
    \end{subfigure}
    \begin{subfigure}[t]{0.24\textwidth}
    \centering
    \includegraphics[width=0.8\textwidth,height=3.5cm]{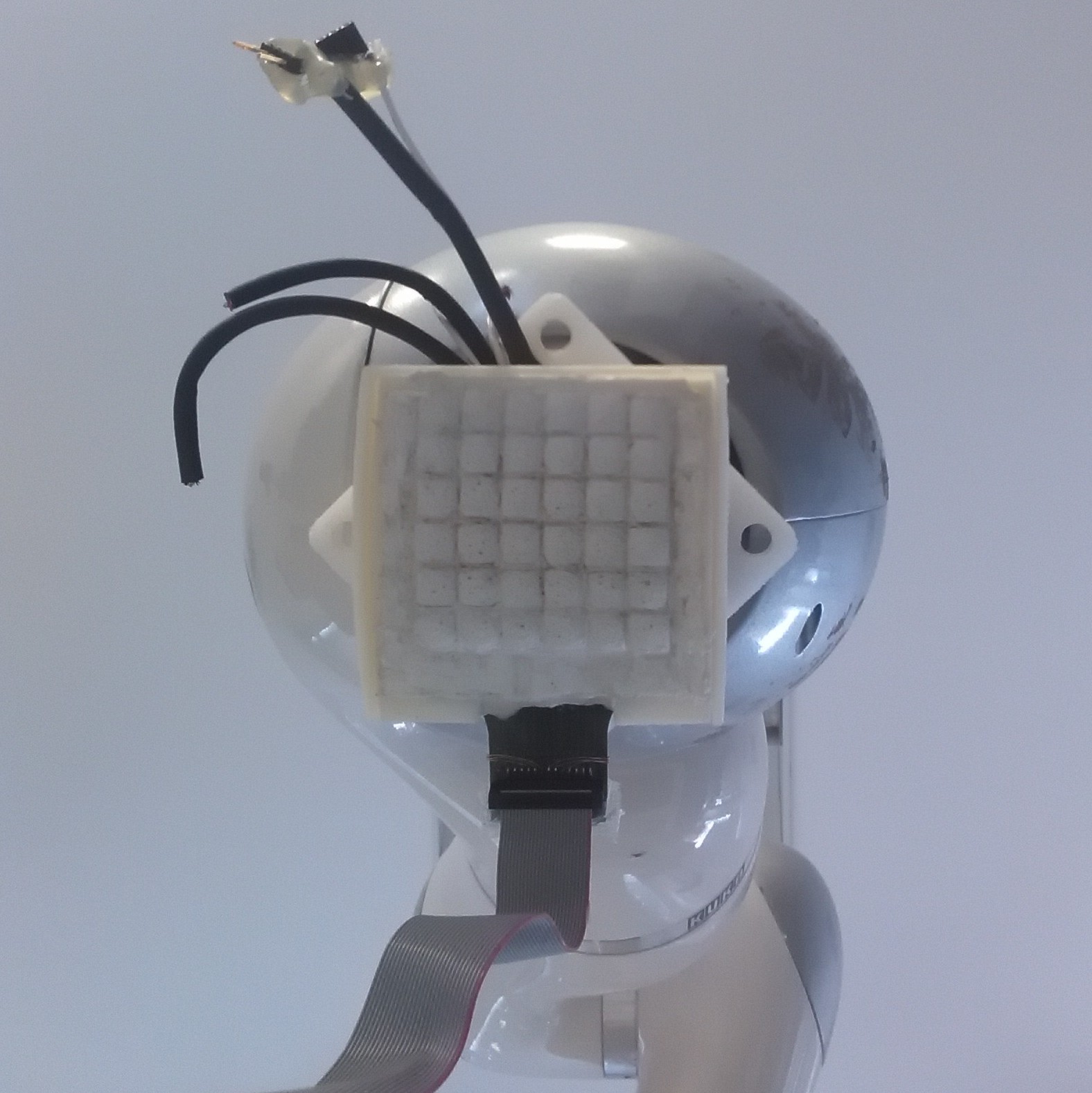}
    \caption{Tactile Skin}
    \label{fig:eeskin}
    \end{subfigure}
     \begin{subfigure}[t]{0.24\textwidth}
    \centering
         \includegraphics[width=0.8\textwidth,height=1.45cm]{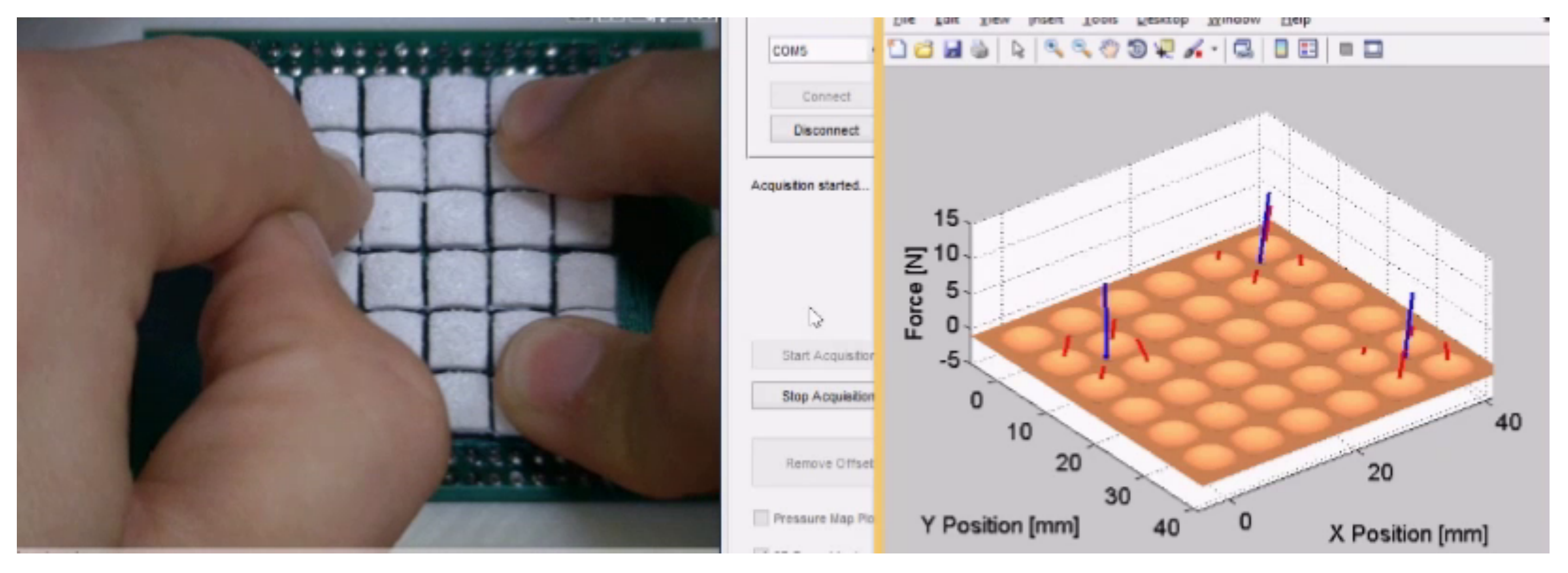}
        \caption{Tactile Skin Readings}
        \label{fig:skinforce}
    \end{subfigure}
         \begin{subfigure}[t]{0.24\textwidth}
        \centering
         \includegraphics[width=0.8\textwidth,height=1.45cm]{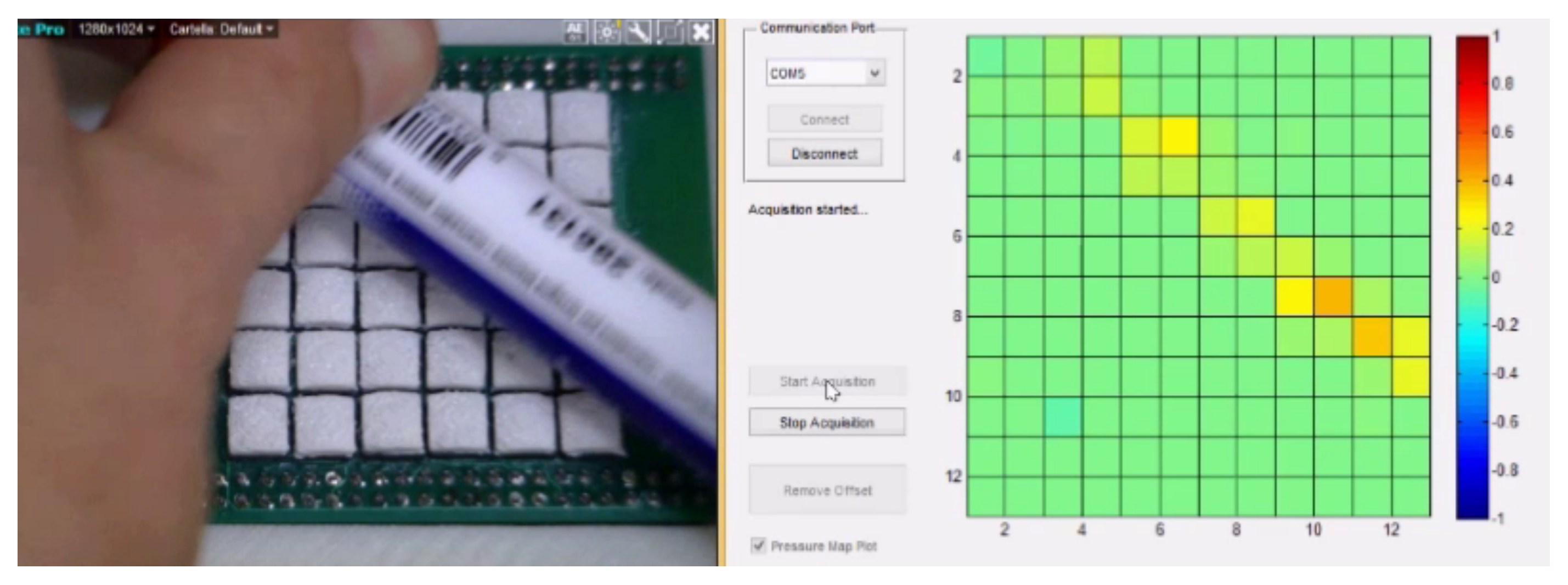}
        \caption{Tactile Skin Readings}
        \label{fig:skintactile}
    \end{subfigure}
    \caption{Experimental setup for tactile perception}
    \label{FIG:tactile_setup}
  \end{figure}

\subsubsection{Exploration Strategy}
For the tactile object representation and recognition, the tactile
exploration is a fundamental phase. An appropriate strategy
guarantees a good quality of the tactile point clouds. For the
experiments discussed in this paper, the objects are explored by
pressing on them along the $\bfz$ axis of the robot base frame,
highlighted in Fig.~\ref{fig:robot}. The base frame has been
selected as the unified world reference frame, and all tactile point
clouds are represented in this frame. During the exploration phase,
if a module of the tactile sensor {senses a contact with the
object $O$, the corresponding point is included in its point cloud
${\cal P}^t$.} The tactile readings are
constituted by six dimensional vector, encoding the pose and the
force data. All exploration experiments have been carried out by
using a Cartesian impedance controller for the robot, in order to
allow a compliant interaction with the objects. In particular, the
robot is compliant along the $z$ axis, so that the exploration
phase does not damage any object. For each object a grid as reported
in Fig.~\ref{FIG:exploration_path} is defined. The end effector of
the robot is moved to each vertex of this grid. Then, after the
reaching of a vertex, the robot presses with the tactile sensor on
the object. The $i$-th point between the skin and the object $\bfp_i$ is
represented by its coordinates $(p_x,p_y,p_z)$, with respect to the
robot base frame. In this paper, the tactile frame $\Sigma_t$
coincides with the robot base frame, but the feature selection is
frame-independent. During the proposed experiments, the objects are
fixed on the table. The points of the table are removed with the
planar filter algorithm implemented in PCL. The simple exploration
strategy, adopted in this paper, is particularly suitable for planar
objects. The whole procedure is detailed in
Algorithm~\ref{a:exploration}.

\begin{algorithm}
    \caption{Exploration Strategy}
    \label{a:exploration}
    \begin{algorithmic} [1]
        \State $Traj_1=(\bfv_1, \bfv_2, ...., \bfv_n)$ \Comment grid vertices in Fig. \ref{FIG:exploration_path}
        \For {$\bfv_j \in Traj_1$}
        \State moveTo($\bfv_j$) \Comment{it brings robot from vertex to vertex}
        \State press on vertex $\bfv_j$
        \For {each sensor module $i$}
        \State read($\bfF_i$)
            \If {$\|\bfF_i\| \geq 0.8N$}
            \State $\bfp_i \gets (p_{x}, p_{y}, p_{z})$
            \State ${\cal P} = {\cal P} $ $\cup$ $ \{\bfp_i\}$
            \EndIf
            \EndFor
            \EndFor
    \end{algorithmic}
\end{algorithm}
\section{Experiments} \label{s:results}
\subsection{Description of the Dataset}
We selected $15$ objects, depicted in Fig. \ref{FIG:objects}, which are typical of domestic and industrial environments. For each object, the tactile exploration procedure, described in Algorithm \ref{a:exploration}  has been repeated 5 times. Then, $40$ samples from each object have been collected with the visual system in Fig. \ref{FIG:visual_setup}. After the data acquisition procedure, we have $40$ visual and $5$ tactile point clouds for each object. The visual point clouds are used to build the a-priori knowledge. In this case study, a-priori knowledge is constituted by the classifier trained with visual data, denoted in our work with the symbol ${\cal M}^v$. The tactile exploration data are then classified exploiting the a priori knowledge ${\cal M}^v$.
It is important to emphasize that, with the proposed approach, the robot can classify an object using the sense of touch  when the object has never been touched before, but only seen by vision. {In this work, we used only rigid object. The application of this strategy to deformable objects is planned as future work.}
\begin{figure}[tb]
	 \begin{subfigure}[b]{0.113\textwidth}
		\centering
		 \includegraphics[width=1\linewidth]{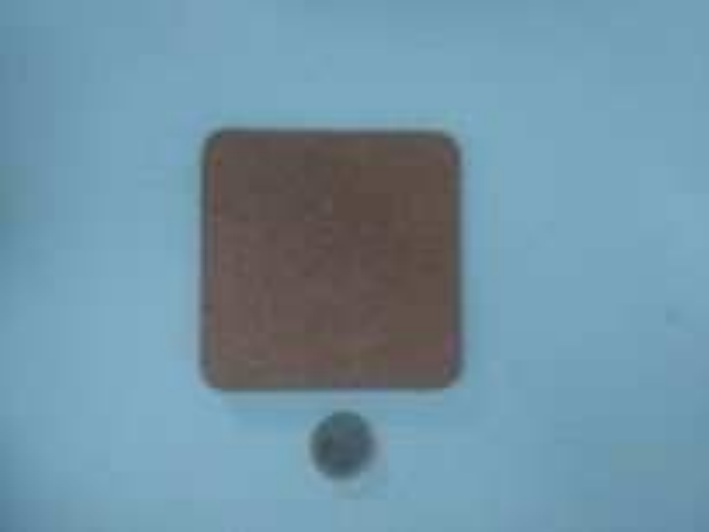}
		\caption{cup mat}
	\end{subfigure}
	 \begin{subfigure}[b]{0.113\textwidth}
		\centering
		 \includegraphics[width=1\textwidth]{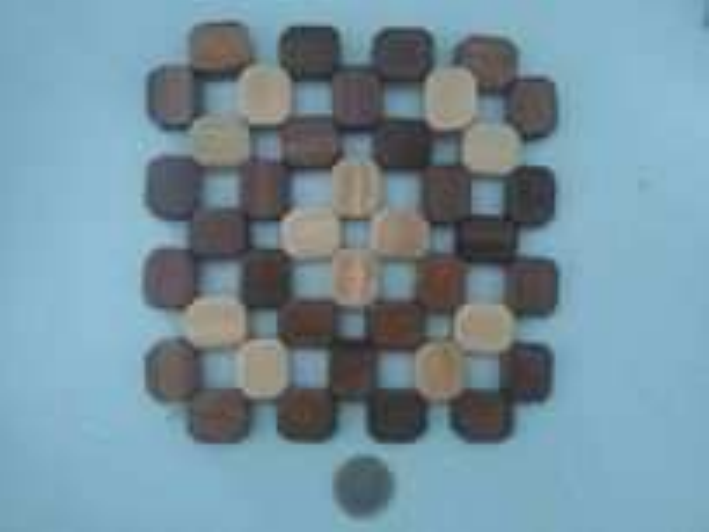}
		\caption{mat}
	\end{subfigure}
	 \begin{subfigure}[b]{0.113\textwidth}
		\centering
		 \includegraphics[width=1\textwidth]{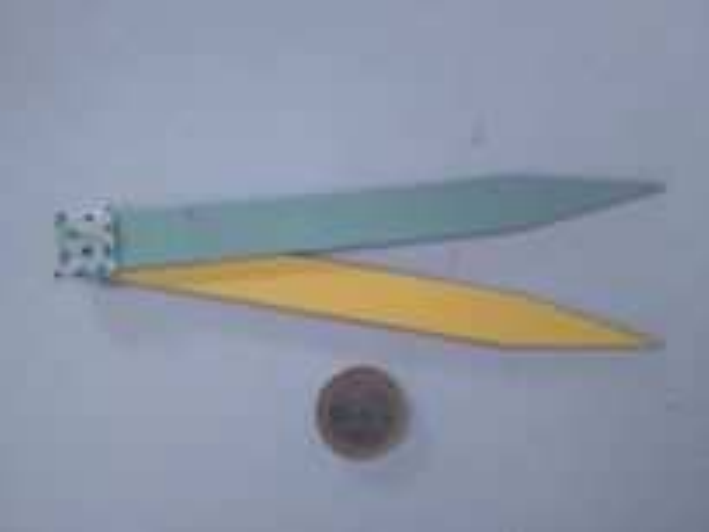}
		\caption{tweezers}
	\end{subfigure}
	 \begin{subfigure}[b]{0.113\textwidth}
		\centering
		 \includegraphics[width=1\textwidth]{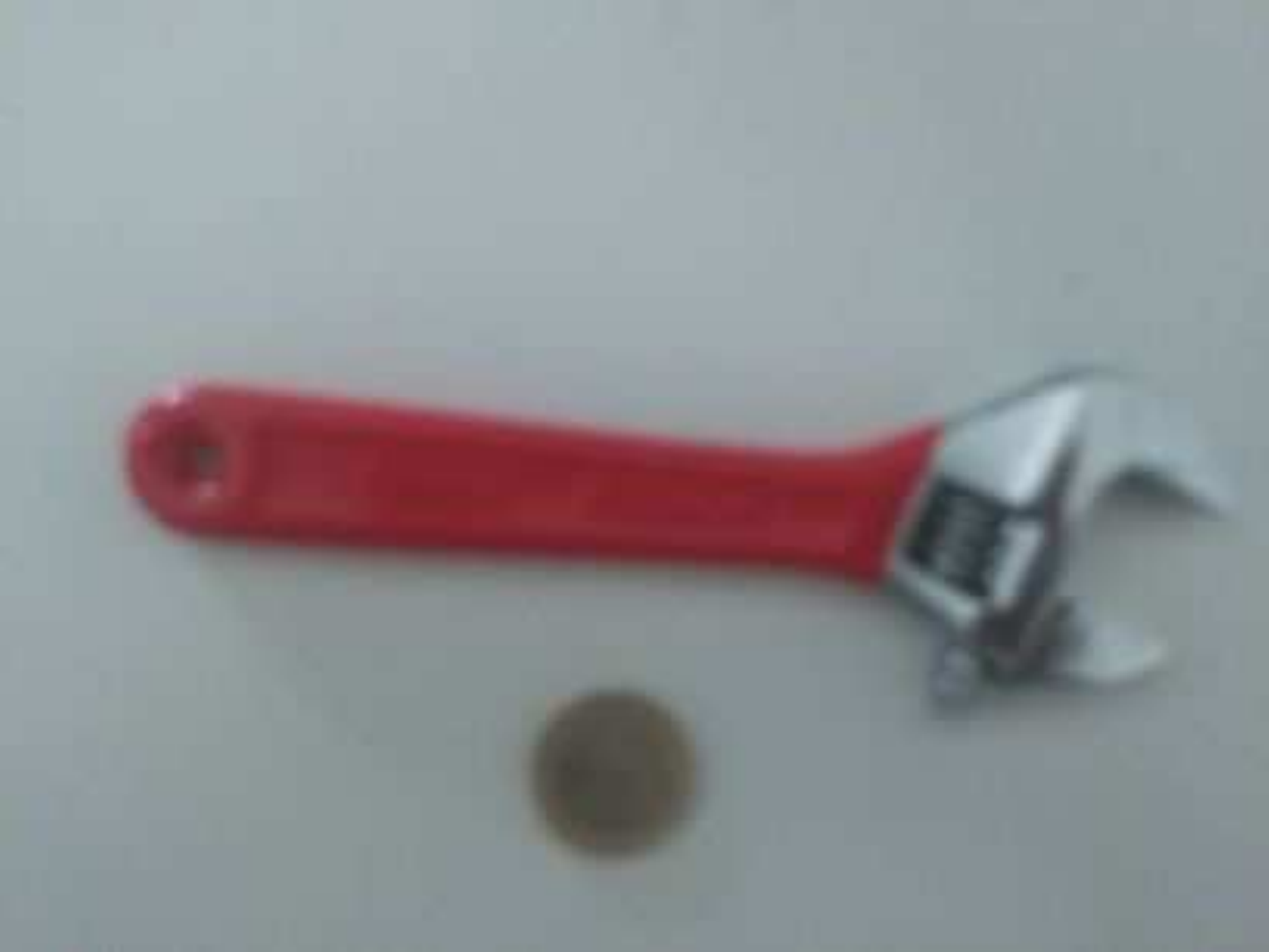}
		\caption{spanner}
	\end{subfigure}
	 \begin{subfigure}[b]{0.113\textwidth}
		\centering
		 \includegraphics[width=1\textwidth]{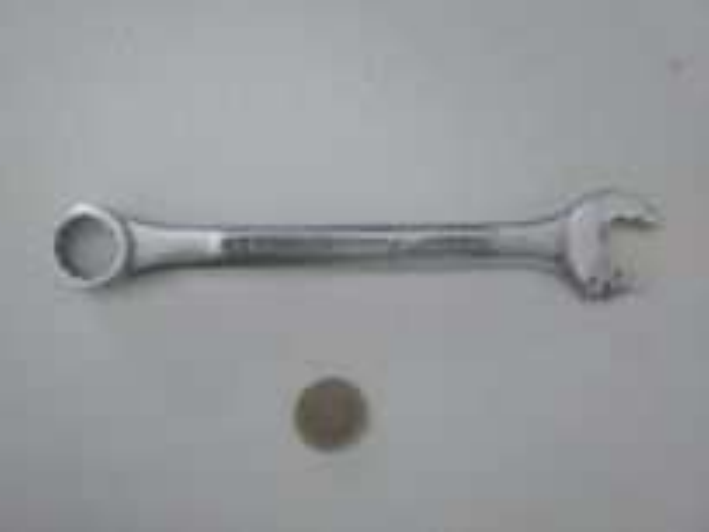}
		\caption{socket wr.}
	\end{subfigure}
	 \begin{subfigure}[b]{0.113\textwidth}
		\centering
		 \includegraphics[width=1\textwidth]{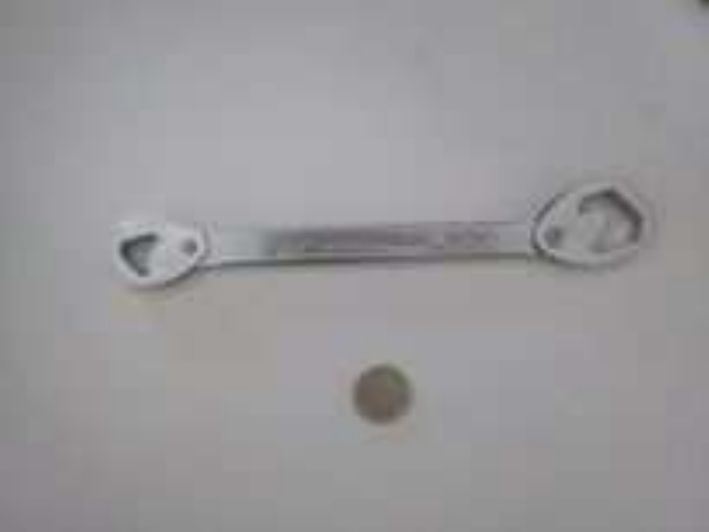}
		\caption{wrench}
	\end{subfigure}
	 \begin{subfigure}[b]{0.113\textwidth}
		\centering
		 \includegraphics[width=1\textwidth]{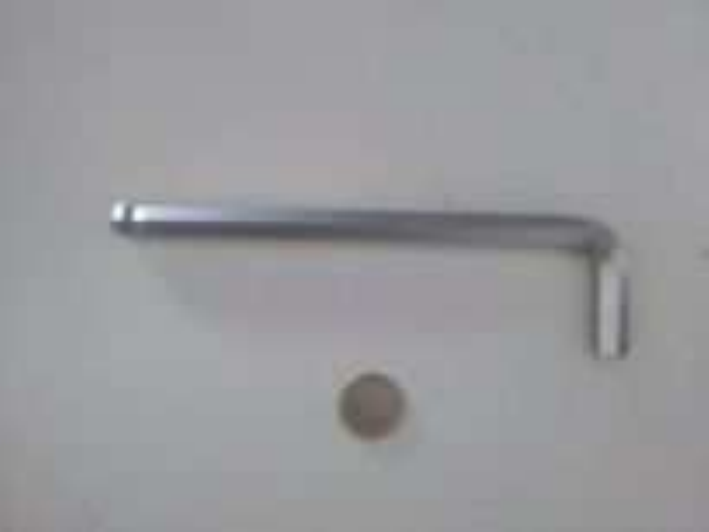}
		\caption{allen key}
	\end{subfigure}
	 \begin{subfigure}[b]{0.113\textwidth}
		\centering
		 \includegraphics[width=1\textwidth]{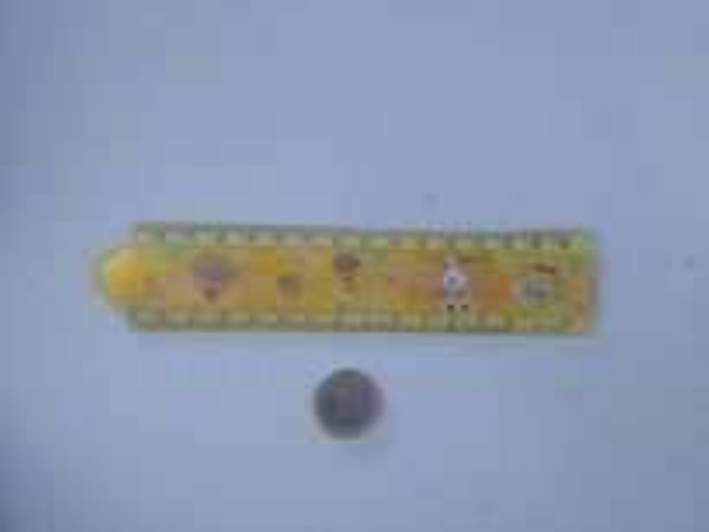}
		\caption{ruler}
	\end{subfigure}
	 \begin{subfigure}[b]{0.113\textwidth}
		\centering
		 \includegraphics[width=1\textwidth]{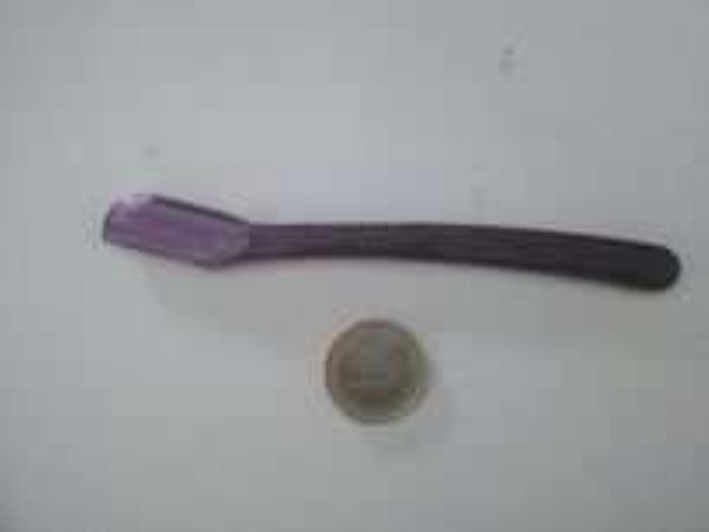}
		\caption{shaver}
	\end{subfigure}
	 \begin{subfigure}[b]{0.113\textwidth}
		\centering
		 \includegraphics[width=1\textwidth]{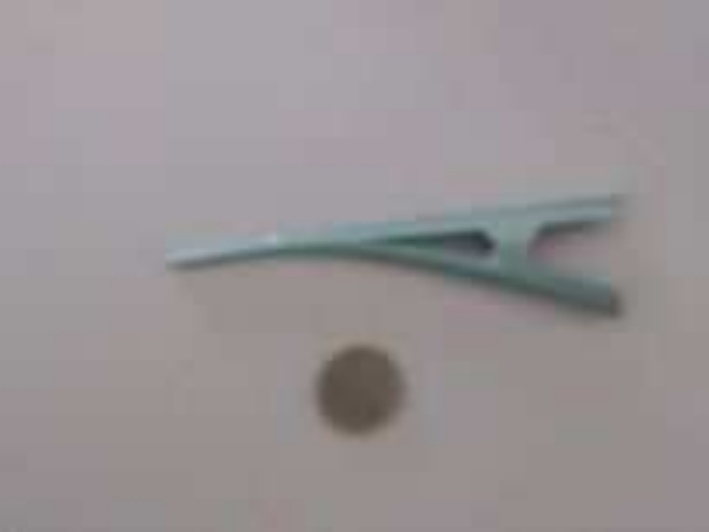}
		\caption{hairpin}
	\end{subfigure}
	 \begin{subfigure}[b]{0.113\textwidth}
		\centering
		 \includegraphics[width=1\textwidth]{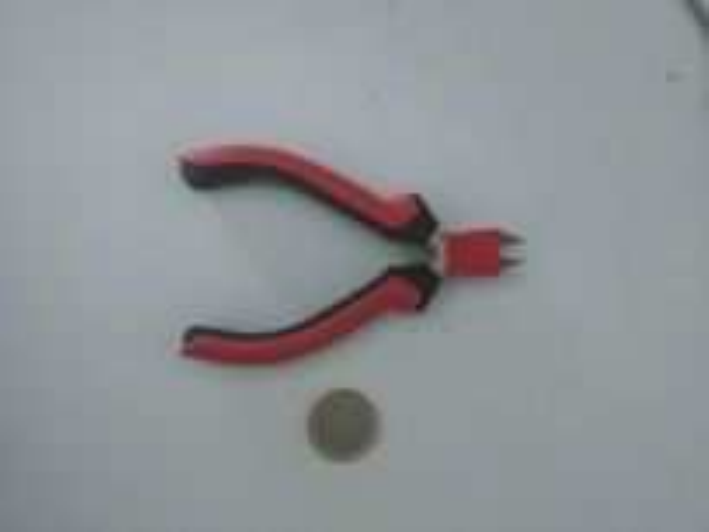}
		\caption{pincers}
	\end{subfigure}
	 \begin{subfigure}[b]{0.113\textwidth}
	\centering
	 \includegraphics[width=1\textwidth]{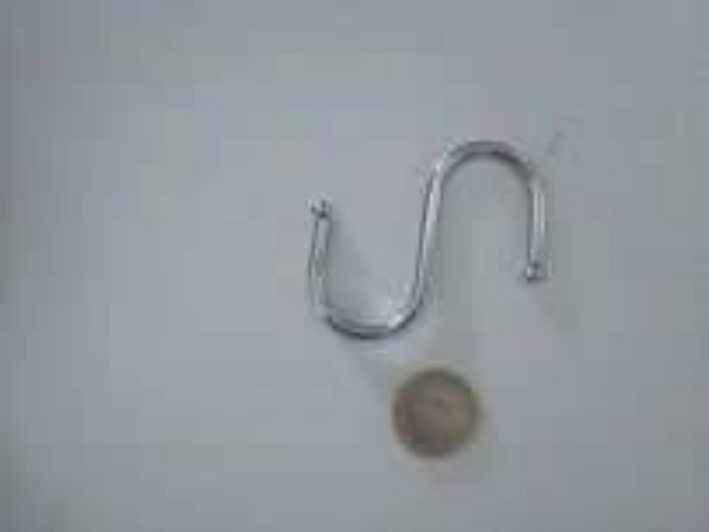}
	\caption{holder}
	\end{subfigure}
	 \begin{subfigure}[b]{0.113\textwidth}
	\centering
	 \includegraphics[width=1\textwidth]{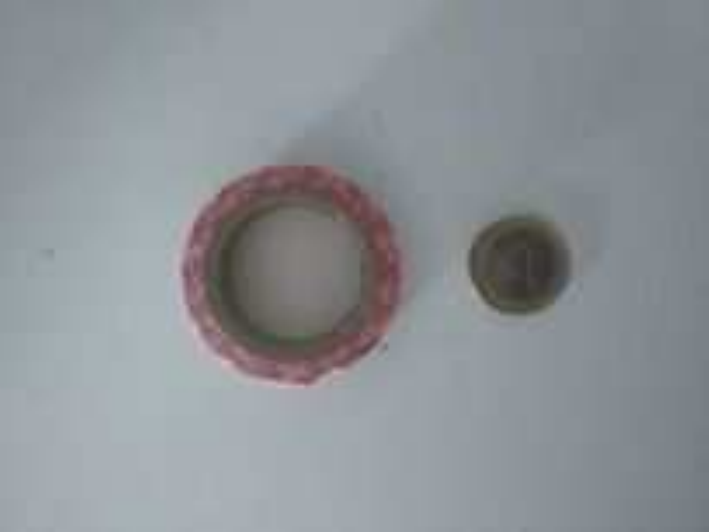}
	\caption{small tape}
	\end{subfigure}
	 \begin{subfigure}[b]{0.113\textwidth}
	\centering
	 \includegraphics[width=1\textwidth]{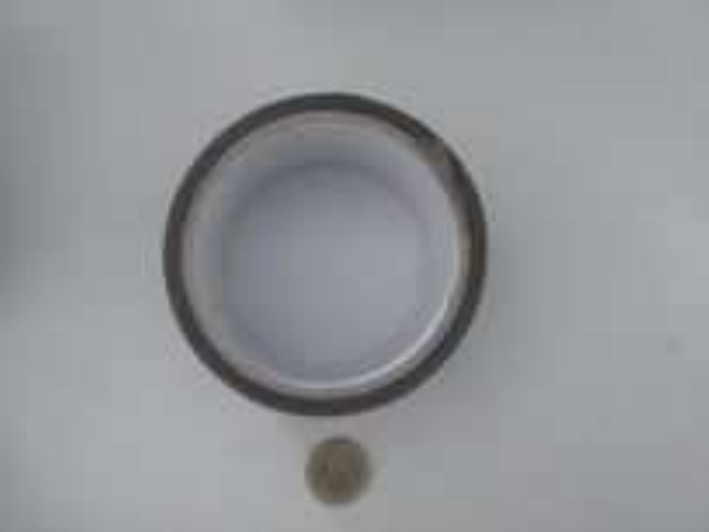}
	\caption{tape}
	\end{subfigure}
	 \begin{subfigure}[b]{0.113\textwidth}
		\centering
		 \includegraphics[width=1\textwidth]{object24.eps}
		\caption{mouse}
	\end{subfigure}
		\caption{{Objects used for experiments, depicted near to a 1-euro coin}}
		\label{FIG:objects}
\end{figure}
\begin{table}[tbh]
	\centering
	\begin{tabular}{l*{5}{c}r}
		Feature descriptor & 1-NN & 3-NN & 5-NN & SVM \\
		\hline
        PFH & $15.20\%$ & $8.44\%$ & $8.44\%$ & $8.44\%$ \\	
        FPFH & $10.06\%$ & $10.06\%$ & $10.06\%$ & $13.34\%$ \\	
        SI & $21.33\%$ & $22.67\%$ & $21.33\%$ & $32.00\%$\\	
		SHOT & $32.63\%$ & $32.63\%$ & $31.25\%$ & $35.79\%$\\
        ESF & $45.26\%$ & $45.26\%$ & $42.11\%$ & $32.63\%$\\
        $\bfd_{c}$ & $45.26\%$ & $45.26\%$ & $42.11\%$ & $26.32\%$\\
		CLUE& $55.79\%$ & $\mathbf{57.89\%}$& $53.68\%$ & $33.68\%$ \\
	\end{tabular}
	\caption{Cross-modal Recognition result without preprocessing}
	\label{t:result_no_preprocessing}
\end{table}
\begin{table}[htbp]
	\centering
	\begin{tabular}{l*{3}{c}r}
		Feature Descriptor & 1-NN & 3-NN &  5-NN &  SVM \\
		\hline
        PFH & $5.33\%$ & $12.00\%$& $12.00\%$ & $10.67\%$ \\
		FPFH & $9.33\%$ & $12.00\%$& $14.67\%$ & $16.00\%$ \\
       	SI & $22.67\%$ & $28.00\%$& $28.00\%$ & $40.00\%$\\
        SHOT & $37.33\%$ & $36.00\%$& $34.67\%$ & $32.00\%$\\
        ESF  & $60.00\%$ & $65.33\%$& $65.33\%$ & $49.33\%$\\
        $\bfd_{c}$ & $62.67\%$ & $68.00\%$ & $66.67\%$ & $30.67\%$\\
        CLUE & $72.00\%$ & $73.33\%$ & $\mathbf{77.33}\%$ & $40.00\%$\\

	\end{tabular}
	\caption{Cross-modal Recognition Result with preprocessing}
	 \label{t:result_with_preprocessing}
\end{table}
\subsection{Classification results}
In order to assess the performance of the framework, we evaluate, in terms of accuracy, the proposed combination of (1) unified representation, (2) unified descriptor, (3) transfer learning approach, and (4) learning algorithm. The accuracy is defined as the number of correct classification operations over the total number of classification operations.   We compare different state-of-the-art descriptors with the proposed CLUE, as shown in Table \ref{t:result_no_preprocessing} and Table \ref{t:result_with_preprocessing}. Also, we show how the application of transfer learning techniques improves the performance in terms of recognition accuracy.

\subsubsection{Results without Transfer Learning}
First, we evaluate the proposed framework without using transfer learning. In this case study, we adopt the Cross Modal Recognition process (CMR) proposed in \cite{falco2017cross}.
In order to show how strongly the preprocessing step described in Sec. \ref{s:unified} impacts on the performance of cross-modal object recognition, we indicate in Table \ref{t:result_no_preprocessing} and Table \ref{t:result_with_preprocessing} the classification accuracy without and with preprocessing, respectively. 
In the light of the results mentioned above, the preprocessing allows {an improvement of up to $20\%$} in the recognition accuracy and the importance of this step is crucial when dealing with visuo-tactile cross-modality. As classification algorithms, we compare $k$-NN and radial basis function kernel SVM, since they are simple and have recognized effectiveness in classification problems. The best performance is achieved by $5$-NN combined with the CLUE descriptor.

The accuracy is evaluated by classifying five tactile explorations per object using the visual knowledge embedded in the model ${\cal M}^v$, obtained through the different classification methods reported in Table \ref{t:result_with_preprocessing}. We can observe that the descriptors using estimated normal vectors, i.e., PFH, FPFH, SI and SHOT, perform worse than ESF.
CLUE is particularly suitable for cross-modal recognition, as the improvement with respect to basic ESF is, in our case study, almost $15\%$ and the dimension remains the same.
CMR achieves $77\%$ accuracy. In Table \ref{t:confusion_CMR}, the confusion matrix is reported for a more detailed analysis. We can notice the recognition performance is between $80\%$ and $100\%$ for all the objects except for {four objects}: {(j) the hairpin achieves $20\%$ accuracy,  (n) the tape $40\%$ accuracy, (e) the socket wrench $60\%$ accuracy, and (h) the ruler $60\%$ accuracy.}
\begin{table}[t]
	\centering
    \resizebox{\columnwidth}{!}{%
	\begin{tabular}{l|l*{15}{c}r|}
		\hline
        a & \cellcolor{gray}1.0 & 0 & 0 & 0 & 0 & 0 & 0 &\cellcolor{gray!20}0.2 & 0 & 0 & 0 & \cellcolor{gray!20}0.2 & 0 & 0 & 0 \\
        b & 0 & \cellcolor{gray}1.0 & 0 & 0 & 0 & 0 & 0 & 0 & 0 & 0 & 0 & 0 & 0 & 0 & 0 \\
		c & 0 & 0 & \cellcolor{gray}1.0 & 0 & \cellcolor{gray!20}0.4 & 0 & 0 & 0 & 0 & 0 & 0 & 0 & 0& 0 & 0 \\
		d & 0 & 0 & 0 & \cellcolor{gray}1.0 & 0 & 0 & 0 & 0 & 0 & 0 & 0 & 0 & 0 & 0 & 0 \\
		e & 0 & 0 & 0 & 0 & \cellcolor{gray}0.6 & 0 & 0 & 0 & 0 & 0 & 0 & 0 & 0 & 0 & 0 \\
		f & 0 & 0 & 0 & 0 & 0 & \cellcolor{gray}0.8 & 0 & 0 & 0 & 0 & 0 & 0 & \cellcolor{gray!20}0.2 & 0 & 0 \\
		g & 0 & 0 & 0 & 0 & 0 & \cellcolor{gray!20}0.2 & \cellcolor{gray}1.0 & 0 & 0 & 0 & 0 & 0 & 0 & 0 & 0 \\
		h & 0 & 0 & 0 & 0 & 0 & 0 & 0 & \cellcolor{gray}0.6 & 0 & 0 & 0 & 0 & 0 & 0 & 0 \\
		i & 0 & 0 & 0 & 0 & 0 & 0 & 0 & \cellcolor{gray!20}0.2 & \cellcolor{gray}1.0 & 0 & 0 & 0 & 0 & \cellcolor{gray!20}0.6 & 0 \\
		j & 0 & 0 & 0 & 0 & 0 & 0 & 0 & 0 & 0 & \cellcolor{gray}0.2 & 0 & 0 & 0 & 0 & 0 \\
		k & 0 & 0 & 0 & 0 & 0 & 0 & 0 & 0 & 0 & \cellcolor{gray!20}0.8 & \cellcolor{gray}1.0 & 0 & 0 & 0 & 0\\
		l & 0 & 0 & 0 & 0 & 0 & 0 & 0 & 0 & 0 & 0 & 0 & \cellcolor{gray}0.8 & 0 & 0 & 0 \\
		m & 0 & 0 & 0 & 0 & 0 & 0 & 0 & 0 & 0 & 0 & 0 & 0 & \cellcolor{gray}0.8 & 0 & 0 \\
		n & 0 & 0 & 0 & 0 & 0 & 0 & 0 & 0 & 0 & 0 & 0 & 0 & 0 & \cellcolor{gray}0.4 & 0 \\
		
		o & 0 & 0 & 0 & 0 & 0 & 0 & 0 & 0 & 0 & 0 & 0 & 0 & 0 & 0 & \cellcolor{gray}1.0 \\
		\hline
		& a & b & c & d & e & f & g & h & i & j & k & l & m & n & o \\
	\end{tabular}}
	\caption{{Confusion matrix for the CMR architecture with CLUE descriptor and 5-NN classification algorithm.}} 
	\label{t:confusion_CMR}
\end{table}
\begin{table}[t]
	\centering
    \resizebox{\columnwidth}{!}{%
	\begin{tabular}{l|l*{15}{c}r|}
		\hline
        a & \cellcolor{gray}1  & 0 & 0 & 0 & 0 & 0 & 0 & 0 & 0 & 0 & 0 & 0 & 0 & 0 & 0 \\
        b & 0 & \cellcolor{gray}1.0 & 0 & 0 & 0 & 0 & 0 & 0 & 0 & 0 & 0 & 0 & 0 & 0 & 0 \\
		c & 0  & 0 & \cellcolor{gray}1  & 0 & 0 & 0 & 0 & 0 & 0 & 0 & 0 & 0 & 0 & 0 & 0 \\
		d & 0 & 0 & 0 & \cellcolor{gray}1.0 & \cellcolor{gray!20}0.2  & 0 & 0 & 0 & 0 & 0 & 0 & 0 & 0 & 0 & 0 \\
		e & 0 & 0 & 0 & 0 & \cellcolor{gray}0.6 & 0 & 0 & 0 & 0 & 0 & 0 & 0 & 0 & 0 & 0 \\
		f & 0 & 0 & 0 & 0 & \cellcolor{gray!20}0.2 & \cellcolor{gray}1.0 & 0 & 0 & 0 & 0 & 0 & 0 & 0 & 0 & 0 \\
		g & 0 & 0 & 0 & 0 & 0 & 0 & \cellcolor{gray}1.0 & 0 & 0 & 0 & 0 & 0 & 0 & 0 & 0 \\
		h & 0 & 0 & 0 & 0 & 0 & 0 & 0 & \cellcolor{gray}1.0 & 0 & 0 & 0 & 0 & 0 & 0 & 0 \\
		i & 0 & 0 & 0 & 0 & 0 & 0 & 0 & 0 & \cellcolor{gray}1.0 & 0 & 0 & 0 & 0 & 0 & 0 \\
		j & 0 & 0 & 0 & 0 & 0 & 0 & 0 & 0 & 0 & \cellcolor{gray}1.0 & 0 & 0 & 0 & 0 & 0 \\
		k & 0 & 0 & 0 & 0 & 0 & 0 & 0 & 0 & 0 & 0 & \cellcolor{gray}1.0 & 0 & 0 & 0 & 0\\
		l & 0 & 0 & 0 & 0 & 0 & 0 & 0 & 0 & 0 & 0 & 0 & \cellcolor{gray}1.0 & 0 & 0 & 0 \\
		m & 0 & 0 & 0 & 0 & 0 & 0 & 0 & 0 & 0& 0 & 0 & 0 & \cellcolor{gray}1.0 & 0 & \cellcolor{gray!20}0.4 \\
		n & 0 & 0 & 0 & 0 & 0 & 0 & 0 & 0 & 0 & 0 & 0 & 0 & 0 & \cellcolor{gray}1.0 & 0 \\
		o & 0 & 0 & 0 & 0 & 0 & 0 & 0 & 0 & 0 & 0 & 0 & 0 & 0 & 0 &  \cellcolor{gray}0.6 \\
		\hline
		& a & b & c & d & e & f & g & h & i & j & k & l & m & n & o \\
	\end{tabular} }
	\caption{ {Confusion matrix for the TL-CMR architecture with CLUE descriptor, GFK transfer learning, and RBF SVM classifier.}} 
	\label{t:confusion_TL-CMR}
\end{table}
\subsubsection{Results with Transfer Learning}
Our second analysis is performed by applying the transfer learning approaches described in Sec. \ref{sec:TL}, {hence, using the Transfer Learning-based} Cross Modal Recognition process (TL-CMR).
We used four transfer learning techniques: the simple PCA, Transfer Component Analysis (TCA), Subspace Alignment (SA), and Geodesic Flow Kernel (GFK). In Table \ref{t:trasnfrer_learning_approach_without_preprocessing__visual_tactile}, we report the classification accuracy without preprocessing, i.e., without the equalization procedure performed in Sec. \ref{s:unified}.
The transfer learning approaches based on dimensionality reduction such as PCA, TCA and SA show poor accuracy.
The best approach is GFK combined with CLUE, which improves the accuracy of CMR, achieving $82\%$ accuracy.
Table \ref{t:trasnfrer_learning_approach_with_preprocessing_visual_tactile} reports the results of TL-CMR {including} the equalization steps. Also in this case, PCA, TCA, and SA do not {significantly improve} the performance with respect to the CMR architecture. In fact, in most transfer learning approaches the accuracy is even lower than the CMR accuracy.
However, GFK combined with the CLUE descriptor achieves $94\%$ accuracy using the RBF-SVM classification algorithm and $89\%$ using a simple $3$-NN. {We can see that the equalization step allows a $13\%$ increase in the accuracy when using the TL-CMR architecture as well.} With respect to the simple CMR architecture, using the GFK transfer learning approaches increase the accuracy of almost $20\%$.
In Table \ref{t:confusion_TL-CMR}, the confusion matrix is reported for the best case of TL-CMR architecture. We can notice that most objects are correctly classified in all the trials, except object o (mouse) and object e (socket wrench). In particular, the mouse is confused in $40\%$ of the cases with the small tape. The socket wrench is confused in $20\%$ of the cases with the wrench and in $20\%$ of cases with the spanner.
For GFK, we choose the parameters $d=27$, following the guideline in \cite{gong2012geodesic} and fine-tuning with a grid search.
While GFK does well, methods based on principal component analysis perform badly and reduce the accuracy with respect to the CMR case. Our guess is that a dimensionality reduction based on the covariance matrix such as PCA is not convenient for the CLUE descriptor, which can contain important information with low variance.
{Even though} GFK uses PCA internally, the final measurement of the distance is performed taking into account all the elements of the descriptors, as shown in Eq. (\ref{eq:quadratic_form}). {As a consequence, GFK does not cut off significant low-variance information.}
In order to give at least a rough clue on the computational time efficiency of the proposed approach, we estimated the time required for classifying the whole tactile dataset, given the knowledge acquired from the visual dataset. The classification time for the entire dataset is $0.6\,$s using CLUE + GFK with $3$-NN implemented in MATLAB, and $0.02\,$s using SVM with a Python implementation. The algorithm was executed on a personal computer with an Intel i7-8550 CPU.
\begin{table}[htbp]
	\centering
	\begin{tabular}{|c|c|c|c|c|}
        \hline
		Feature Descriptor & 1-NN  & 3-NN &  Linear SVM  & RBF SVM \\
		\hline
        SHOT + PCA  & $49.3\%$ & $48.0\%$ & $37.3\%$& $41.3\%$ \\
		ESF + PCA  & $53.3\%$ & $58.6\%$& $46.4\%$ & $57.3\%$ \\
       	CLUE + PCA  & $42.6\%$ & $46.7\%$& $57.3\%$ & $57.3\%$\\
        SHOT + TCA  & $25.3\%$ & $18.6\%$& $22.6\%$ & $44.0\%$ \\
		ESF + TCA  & $54.6\%$ & $64.0\%$& $40.0\%$ & $60.0\%$ \\
       	CLUE + TCA  & $44.0\%$ & $40.0\%$& $54.6\%$ & $57.3\%$\\
        SHOT + SA  & $37.3\%$ & $33.3\%$& $24.0\%$ & $33.3\%$ \\
		ESF + SA  & $45.3\%$ & $50.6\%$& $32.0\%$ & $52.0\%$ \\
       	CLUE + SA  & $58.6\%$ & $56.0\%$& $36.0\%$ & $40.0\%$\\
        SHOT + GFK & $50.6\%$ & $50.6\%$& $29.3\%$ & $53.3\%$\\
        ESF + GFK  & $68.0\%$ & $65.3\%$& $54.6\%$ & $74.6\%$\\
        \textbf{CLUE + GFK} & $\mathbf{82.6\%}$ & $\mathbf{82.6\%}$ & $\mathbf{58.6\%}$ & $\mathbf{81.3\%}$ \\
        \hline
	\end{tabular}
	\caption{Cross-modal recognition result applying transfer learning without preprocessing. Visual data are used as a training set, tactile data are used as a test set.}
	 \label{t:trasnfrer_learning_approach_without_preprocessing__visual_tactile}
\end{table}
\begin{table}[htbp]
	\centering
	\begin{tabular}{|c|c|c|c|c|}
        \hline
		Feature Descriptor & 1-NN  & 3-NN &  Linear SVM  & RBF SVM \\
		\hline
        SHOT + PCA  & $53.3\%$ & $53.3\%$ & $48.0\%$& $65.3\%$ \\
		ESF + PCA  & $54.8\%$ & $48.0\%$& $56.0\%$ & $62.7\%$ \\
       	CLUE + PCA  & $52.0\%$ & $46.7\%$& $41.3\%$ & $53.3\%$\\
        SHOT + TCA  & $41.3\%$ & $42.7\%$& $17.3\%$ & $38.7\%$ \\
		ESF + TCA  & $41.3\%$ & $49.3\%$& $52.0\%$ & $57.3\%$ \\
       	CLUE + TCA  & $49.3\%$ & $49.3\%$& $40.0\%$ & $52.0\%$\\
        SHOT + SA  & $33.3\%$ & $34.7\%$& $25.3\%$ & $24.0\%$ \\
		ESF + SA  & $42.7\%$ & $42.7\%$& $14.6\%$ & $48.0\%$ \\
       	CLUE + SA  & $42.7\%$ & $44.0\%$& $34.7\%$ & $48.0\%$\\
        SHOT + GFK & $54.7\%$ & $54.7\%$& $42.7\%$ & $66.7\%$\\
        ESF + GFK  & $77.3\%$ & $81.3\%$& $57.3\%$ & $77.3\%$\\
        \textbf{CLUE + GFK} & $\mathbf{88.0\%}$ & $\mathbf{89.3\%}$ & $\mathbf{76.0\%}$ & $\mathbf{94.7\%}$ \\
        \hline
	\end{tabular}
	\caption{Cross-modal recognition result applying transfer learning with preprocessing. Visual data are used as a training set, tactile data are used as a test set.}
	 \label{t:trasnfrer_learning_approach_with_preprocessing_visual_tactile}
\end{table}

\begin{figure}[tb]
\centering
\includegraphics[width=0.25\textwidth]{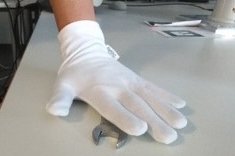}
\caption{Tactile exploration performed by a human}
\label{FIG:human_recogtion}
\end{figure}
\subsection{Comparison with human cross-modal object recognition}
In order to have an ideal reference for assessing the performance of artificial cross-modal recognition, and because in the literature it is hard to find a cross-modal recognition algorithm, we compare the performance of our framework with a "golden standard", which is represented by the performance of humans.  { It is important to note that a rigorous, complete comparison with human performance is beyond the scope of the paper. The comparison with humans is intended to provide a rough indicator of reasonable cross-modal recognition performance.}
We arranged a simple experiment, described in Algorithm \ref{a:protocol}, to have a first estimation of human performance in visuo-tactile cross-modal object recognition tasks.
In this experiment, $10$ participants, {ranging in age from 20 to 30 years}, were invited to look at the set of objects shown in Figure \ref{FIG:objects} for $2$ minutes. Afterwards, each participant was blindfolded and explored each object with one hand. Since human skin can sense also the temperature of the object, the participants wore a thin glove to maintain the tactile perception capability, but to reduce the perception of the temperature. The objects were placed on the table.
The participants explored by touching each object
for $10\,$s without seeing the objects. After that, the participants removed the blindfold and were invited to say which object was explored by looking at the objects. This experimental protocol is summarized in Algorithm \ref{a:protocol}. A picture of human tactile exploration is shown in Fig. \ref{FIG:human_recogtion}. The average accuracy achieved is $89.7\%$ and in Table \ref{t:confusion_human} the confusion matrix is reported. The accuracy  of humans in this experiment is $12\%$ better than the performance of the CMR method based on processed tactile point clouds and the CLUE descriptor.
However, using the TL-CMR method the performance of the robot becomes $6\%$ better than the human estimated performance.
Comparing Tables \ref{t:confusion_CMR} with \ref{t:confusion_human}, we can notice that for most objects the performance of CMR are close to human performance in this case study. It is important to note that the tactile skin does not measure the roughness since it only touches the object without rubbing.
\begin{algorithm}[tb]
	\caption{Protocol for the experiment with humans}
	\label{a:protocol}
	\begin{algorithmic} [1]
    \State The subject is invited to see all the objects for $2$ minutes
    \State The subject is invited to wear a thin glove that prevents from sensing the temperature of  objects
    \State The subject is blinded with a blind fold \label{line:step1}
	\State {The objects are put in a bag, one object is picked out and put on a table}
	\State {The subject explores the object with the hand for $10\,$s}
    \State {The subject removes the blindfold}
    \State The subject sees all the objects and tells the name of the explored object
    \State Go to step \ref{line:step1} until every object has been picked out
\end{algorithmic}
\end{algorithm}
\begin{table}[t]
	\centering
    \resizebox{\columnwidth}{!}{%
	\begin{tabular}{l|l*{15}{c}r|}
		\hline
        a & \cellcolor{gray}0.95  & 0 & 0 & 0 & 0 & 0 & 0 & 0 & 0 & 0 & 0 & 0 & 0 & 0 & 0 \\
        b & 0 & \cellcolor{gray}1.0 & 0 & 0 & 0 & 0 & 0 & 0 & 0 & 0 & 0 & 0 & 0 & 0 & 0 \\
		c & \cellcolor{gray!20}0.05  & 0 & \cellcolor{gray}0.75  & 0 & 0 & 0 & 0 & 0 & 0 & 0 & 0 & 0 & & 0 & 0 \\
		d & 0 & 0 & \cellcolor{gray!20}0.05 & \cellcolor{gray}0.75 & \cellcolor{gray!20}0.1  & 0 & 0 & 0 & 0 & \cellcolor{gray!20}0.1 & 0 & 0 & 0 & 0 & 0 \\
		e & 0 & 0 & 0 & \cellcolor{gray!20}0.15 & \cellcolor{gray}0.9 & \cellcolor{gray!20}0.1 & 0 & 0 & 0 & 0 & 0 & 0 & 0 & 0 & 0 \\
		f & 0 & 0 & 0 & 0 & 0 & \cellcolor{gray}0.9 & 0 & 0 & 0 & 0 & 0 & 0 & 0 & 0 & 0 \\
		g & 0 & 0 & 0 & 0 & 0 & 0 & \cellcolor{gray}0.95 & 0 & 0 & 0 & 0 & 0 & 0 & 0 & 0 \\
		h & 0 & 0 & \cellcolor{gray!20}0.1 & 0 & 0 & 0 & \cellcolor{gray!20}0.05 & \cellcolor{gray}0.8 & \cellcolor{gray!20}0.2 & 0 & 0 & 0 & 0 & 0 & 0 \\
		i & 0 & 0 & \cellcolor{gray!20}0.1 & 0 & 0 & 0 & 0 & \cellcolor{gray!20}0.2 & \cellcolor{gray}0.8 & 0 & 0 & 0 & 0 & 0 & 0 \\
		j & 0 & 0 & 0 & 0 & 0 & 0 & 0 & 0 & 0 & \cellcolor{gray}0.75 & 0 & 0 & 0 & 0 & 0 \\
		k & 0 & 0 & 0 & \cellcolor{gray!20}0.1 & 0 & 0 & 0 & 0 & 0 & \cellcolor{gray!20}0.15 & \cellcolor{gray}1.0 & 0 & 0 & 0 & 0\\
		l & 0 & 0 & 0 & 0 & 0 & 0 & 0 & 0 & 0 & 0 & 0 & \cellcolor{gray}1.0 & 0 & 0 & 0 \\
		m & 0 & 0 & 0 & 0 & 0 & 0 & 0 & 0 & 0& 0 & 0 & 0 & \cellcolor{gray}0.95 & \cellcolor{gray!20}0.05 & 0 \\
		n & 0 & 0 & 0 & 0 & 0 & 0 & 0 & 0 & 0 & 0 & 0 & 0 & \cellcolor{gray!20}0.05 & \cellcolor{gray}0.95 & 0 \\
		o & 0 & 0 & 0 & 0 & 0 & 0 & 0 & 0 & 0 & 0 & 0 & 0 & 0 & 0 &  \cellcolor{gray}1.0 \\
		\hline
		& a & b & c & d & e & f & g & h & i & j & k & l & m & n & o \\
	\end{tabular} }
	\caption{Confusion matrix of human object recognition.} 
	\label{t:confusion_human}
\end{table}
{
\subsection{Comparison with monomodal object recognition}
The results of the cross-modal visuo-tactile object recognition framework are also compared with the monomodal visual and monomodal tactile recognition case. The results of the visual and tactile monomodal cases are reported in Table \ref{t:result_monomodal}.
In this case study, the classifier is trained and tested with the same modality. The accuracy has been evaluated with a 10-fold cross-validation method. From Table \ref{t:result_monomodal} it is possible to see that both visual and tactile monomodal problems are, as expected, less challenging than the cross-modal case, since training set and test set are generated from the same perception modality. Most state-of-the-art descriptors achieve more than $90\%$ accuracy in the monomodal case with 1-NN. We also notice that the performance of tactile classification is slightly less accurate than visual classification.
}
\begin{table}[b]
	\centering
	\begin{tabular}{|c|c|c|}
		\hline
		Feature Descriptor & Visual & Tactile \\
		\hline
        SHOT & $97.17\%$ & $92.00\%$ \\
        ESF  & $97.33\%$ & ${94.67}\%$  \\
        CLUE & $\textbf{98.67\%}$ & $94.67\%$ \\
        \hline
	\end{tabular}
\caption{Monomodal Recognition Result}
	\label{t:result_monomodal}
\end{table}

\section{CONCLUSION AND FUTURE WORK} \label{s:conclusion}
In this work, we deal with robotic cross-modal visuo-tactile object recognition. We train a classifier by using visual data from {an Asus Xtion Pro Live} camera and we recognize objects at execution time only with tactile data, without any a-priori tactile information.
The preliminary version of this work \cite{falco2017cross} showed for the first time that cross-modal visuo-tactile object recognition is feasible with respectable performance. It leverages empirical methods such as equalization of partiality and resolution, as well as a novel descriptor that performs well across different modalities.
In this paper, we extend the framework proposed in \cite{falco2017cross} by combining the empirical ideas with formal transfer learning techniques.
We show that combining the equalization of partiality and resolution, the CLUE descriptor, and a transfer learning techniques called geodesic flow kernel, we achieve an accuracy that is very close to the monomodal case.
It is interesting to emphasize that our method reaches the peak of performance only when the transfer learning algorithms
are combined with the pipeline proposed in \cite{falco2017cross}.
Therefore, the mere application of TL without a preprocessing and without using CLUE cannot substitute the {CMR effectively}, but a smart combination significantly improves the results.

Future work will follow different directions. The first is implementing more complex exploration strategies. In this work, we used a simple strategy suitable mainly for quasi-planar rigid objects. In order to extend the TL-CMR to arbitrary objects, we have to introduce more sophisticated exploration algorithms, based for example on a nonprehensile manipulation strategy. 
{Also, in current implementation the robot has to perform an extensive tactile exploration to build the point cloud. Novel exploration strategies can exploit the measurement of both tangential and normal contact forces to achieve a more effective exploration by manipulating the object and selecting specific sample points.}
A second direction is investigating novel visuo-tactile descriptors or in applying deep learning methods to train a model from a huge amount of visual data and transfer the acquired knowledge to a small amount of tactile data. Collecting a huge amount of tactile data, in fact, can be unpractical as robot and environments are in physical contact. {In fact, robot movements for collecting data have a cost in terms of energy and tactile sensors can deteriorate, thus making collected data less accurate.} On the other hand, visual images are less expensive to collect and, therefore, more suitable for approaches based on big data. A future work direction will be therefore to train big deep networks with visual data and reuse the knowledge with tactile data, without the need of huge tactile data collection. {A starting point for our investigation will be 3D Convolutional Neural Networks (CNN) for object recognition \cite{maturana2015voxnet, huang2016point}}.
The third direction is to perform a larger number of experiments with human subjects {to compare the performance of human and robot more accurately.} Moreover, cross-modal perception will be applied in other fields, such as data efficient learning \cite{saveriano2017learning, falco2018} and object tracking \cite{koo2014incremental}.

\section*{Acknowledgment}
This work has been partially funded by the Marie Curie Action LEACON, EU project 659265.

\ifCLASSOPTIONcaptionsoff
  \newpage
\fi


\vfill
\vfill
\vfill
\begin{IEEEbiography}[{\includegraphics[width=1in,height=1.25in,clip,keepaspectratio]{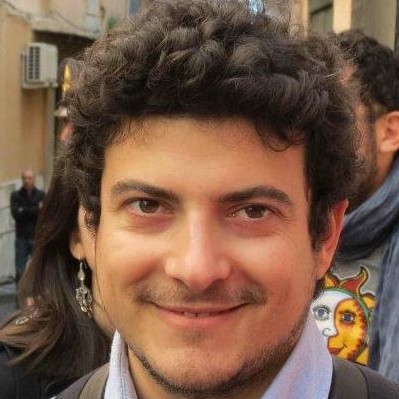}}]
{Pietro Falco}
received his PhD in 2012 at University of Campania Luigi Vanvitelli. From December 2010 to July 2011 he was a visiting scholar at Karlsruhe Institute of Technology (KIT). 
Until 2015, he worked within several EU projects such as DEXMART, ECHORD, SAPHARI and EUROC at University of Campania.
In 2011, Pietro co-founded Aeromechs srl, a successful startup in the field of energy management for aeronautics and home automation.
In 2015, he left the active co-leadership of Aeromechs and moved to Technical University of Munich, in order to continue research and teaching activities in robotics. He was awarded a TUM Foundation Fellowship sponsored by Rohde \& Schwarz in March 2015 and a Marie Curie Individual Fellowship for experienced researchers in March 2016 with the project LEACON "LEArning-CONtrol tight interaction: a novel approach to robust execution of mobile manipulation tasks". From January 2018, Pietro is a tenured senior scientist at ABB corporate research, Sweden.  His research focuses on control theory and machine learning for robotics, human-robot interaction, and human motion interpretation.
\end{IEEEbiography}
\begin{IEEEbiography}[{\includegraphics[width=1in,height=1.25in,clip,keepaspectratio]{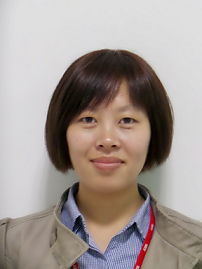}}]{Shuang Lu}
was born in Heilongjiang, China, in 1987. She received the B.E. degree in electrical engineering from Northeast Electric Power University, China, in 2009, and M.Sc. degree in electrical engineering and information technology from Technical University of Munich, Germany, in 2016.
\end{IEEEbiography}
\begin{IEEEbiography}[{\includegraphics[width=1in,height=1.25in,clip,keepaspectratio]{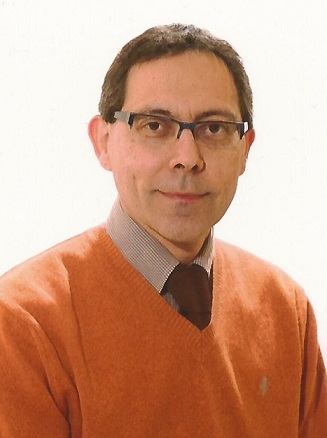}}]{Ciro Natale}
Ciro Natale received the Laurea degree and the Research Doctorate
degree in Electronic Engineering from the University of Naples in
1995 and 2000, respectively. From 2000 to 2004 he has been Research
Associate at the Department of Industrial and Information
Engineering of the University of Campania ``Luigi Vanvitelli'',
where he currently holds the position of Full Professor of
Robotics and Mechatronics. From November 1998 to April 1999 he was a
Visiting Scholar at the German Aerospace Center (DLR) in
Oberpfaffenhofen, Germany. His research interests include modeling
and control of industrial manipulators, force and visual control,
cooperative robots, as well as modeling and control of flexible
structures, active noise and vibration control and modeling,
identification and control of smart actuators. He has published more
than 120 journal and conference papers, including two monographs
published by Springer. He served or is serving as Associate Editor
of various international journals and he had responsibility roles in
many European projects funded under the 7th and the 8th Framework
Programmes.
\end{IEEEbiography}
\vfill
\vfill
\vfill

\begin{IEEEbiography}[{\includegraphics[width=1in,height=1.25in,clip,keepaspectratio]{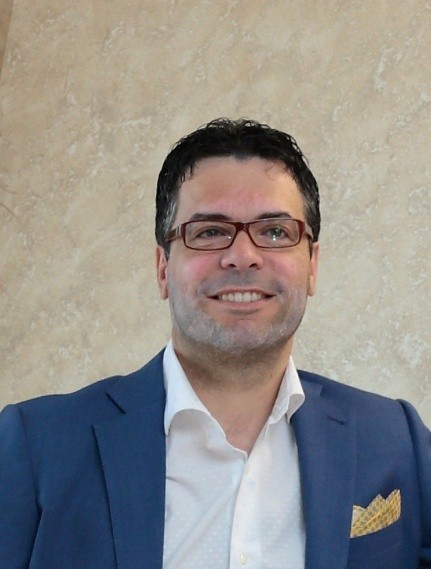}}]{Salvatore Pirozzi}
was born in Napoli, Italy, on April 1977. He received the Laurea and the Research Doctorate degree in Electronic Engineering from the Seconda Universit\`a degli Studi di Napoli, Aversa, Italy, in 2001 and 2004, respectively. From 2008 to 2016 he has been Research Associate at Seconda Universit\`a degli Studi di Napoli. He currently holds the position of Associate Professor at Universit\`a degli Studi della Campania "Luigi Vanvitelli". His research interests, in the aeronautics application sector, include modeling and control of smart actuators for active noise and vibration control. His research activities on robotics include design and modelling of innovative sensors, in particular of tactile solutions, as well as interpretation and fusion of data acquired from the developed sensors. He published more than 70 international journal and conference papers and he is co-author of the book ''Active Control of Flexible Structures'', published by Springer. He currently serves as Associate Editor of the IEEE Transaction on Control Systems Technology.
\end{IEEEbiography}
\begin{IEEEbiography}[{\includegraphics[width=1in,height=1.25in,clip,keepaspectratio]{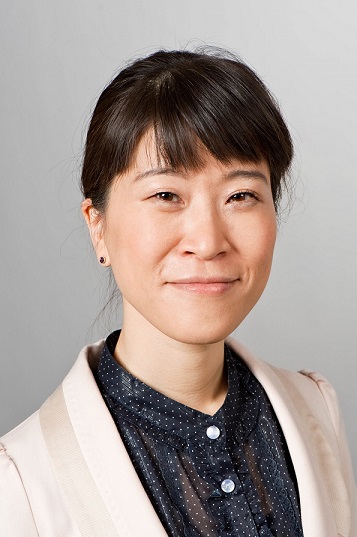}}]
{Dongheui Lee}
is an associate professor at Department of Electrical and Computer Engineering, Technical University of Munich (TUM) and a head of human-centered assistive robotics group at DLR, Germany. Prior, she was assistant professor at TUM (2009-2017), project assistant professor at the University of Tokyo (2007-2009), and a research scientist at Korea Institute of Science and Technology (KIST) (2001-2004). She received mechanical engineering B.S. (2001) and M.S. (2003) degrees from Kyunghee University, Korea and PhD degree (2007) from the department of Mechano-Informatics, the University of Tokyo, Japan. She was awarded for a Carl-von-Linde fellowship at TUM Institute for Advanced Study (2011) and a Helmholtz professorship prize (2015). She is Coordinator of euRobotics Topic Group on physical Human Robot Interaction and the co-coordinator of TUM Center of Competence Robotics, Autonomy and Interaction. Her research interests include human motion understanding, human robot interaction, machine learning in robotics, and assistive robotics.
\end{IEEEbiography}
\vfill
\vfill
\vfill
\vfill


\end{document}